\documentclass{article}

\usepackage[preprint]{neurips_2026}

\usepackage[utf8]{inputenc} 
\usepackage[T1]{fontenc}    
\usepackage{hyperref}       
\usepackage{url}            
\usepackage{booktabs}       
\usepackage{amsfonts}       
\usepackage{nicefrac}       
\usepackage{microtype}      
\usepackage{xcolor}         

\usepackage{amsmath}
\usepackage{amssymb}
\usepackage{mathtools}
\usepackage{amsthm}

\theoremstyle{plain}

\theoremstyle{definition}

\theoremstyle{remark}

\title{Reward Inflation: A Healthy Stimulus for Reinforcement Learning}

\author{
  Ganghun Lee\textsuperscript{1} \quad
  Minji Kim\textsuperscript{2} \quad
  Minsu Lee\textsuperscript{3}\thanks{Corresponding authors.} \quad
  Byoung-Tak Zhang\textsuperscript{1,2}\footnotemark[1] \\
  \textsuperscript{1}AIIS, Seoul National University \\
  \textsuperscript{2}Department of Computer Science, Seoul National University \\
  \textsuperscript{3}School of AI Convergence, Sungshin Women's University \\
  \texttt{\{zxc8594, cyqkcy01\}@snu.ac.kr},
  \texttt{mslee@sungshin.ac.kr},
  \texttt{btzhang@snu.ac.kr}
}

\begin{document}

\maketitle

\begin{abstract}
Reward serves as the primary learning signal in reinforcement learning (RL).
However, while reward magnitudes are typically held fixed throughout training, their temporal modulation remains underexplored.
In this paper, we propose \textit{reward inflation}, a gradual scaling of rewards over the course of training, and show that it can act as a healthy stimulus for RL.
Theoretically, reward inflation induces an implicit recency weighting that upweights recent transitions during policy updates, enabling faster adaptation.
We further show that reward inflation suppresses the emergence of dormant neurons by sustaining gradient signals as the policy saturates, suggesting a possible contribution to plasticity preservation.
Empirical results on ALE games and MuJoCo tasks corroborate these findings, showing that an appropriate level of reward inflation benefits a broad range of tasks.
Finally, we introduce \textit{Fed}, an adaptive variant that adjusts the inflation level on the fly, and find that it often improves upon fixed inflation. Source code is available at \url{https://github.com/northadventure/RewardInflation}.
\end{abstract}

\section{Introduction}

Reinforcement Learning (RL) has shown promising results in diverse domains, such as robotics~\citep{tang2025deep}, games~\citep{souchleris2023reinforcement}, art~\citep{lee2023neural}, autonomous driving~\citep{zhao2024survey}, recommendation systems~\citep{afsar2022reinforcement}, finance~\citep{bai2025review}, and gaining traction to finetune and improve large language models (LLMs)~\citep{liu2025reinforcement}. 
Since RL learns from rewards, optimizing the reward function is central to RL research.
Although reward design is largely task-specific, some previous works have explored various techniques to enhance or augment fully or partially task-agnostic reward design~\citep{eschmann2021reward}, such as reward normalization~\citep{schaul2021return}, potential-based reward shaping~\citep{ng1999policy}, adaptive reward shaping~\citep{ma2024highly}, intrinsic motivation~\citep{aubret2023information}, curriculum rewards~\citep{gupta2022extending}, and, more recently, human- or LLM-augmented rewards~\citep{yu2025rewardmodelsdeepreinforcement}.
However, the temporal dynamics of the reward scale throughout training remain underexplored.

In this paper, we introduce \textit{reward inflation}, a novel approach that scales rewards slowly yet exponentially over the course of RL.
The idea is simple and widely applicable to implement: maintaining a positive variable called the inflation level, which is periodically updated by an inflation rate and multiplied with the reward.
The agent then learns from the inflated rewards during training, while previously stored rewards in the replay buffer remain unchanged.
Additionally, we introduce an optional component, \textit{Fed}, which adaptively adjusts the inflation rate during training by monitoring whether the moving average of the policy gradient norm becomes excessive or insufficient.

Despite its simplicity, reward inflation has rich theoretical implications.
Based on our theoretical analysis under a fixed inflation rate, reward inflation induces an implicit recency weighting in the TD update—a non-uniform weighting across the replay buffer that assigns greater effective weight to recent transitions.
In addition, reward inflation suppresses the emergence of dormant neurons: by gradually increasing the overall gradient norm, it may help preserve plasticity, potentially contributing to improved learning.

Our experimental results on ALE~\citep{bellemare2013arcade} games and MuJoCo~\citep{todorov2012mujoco} tasks show that, under a moderate inflation rate, reward inflation benefits a broad range of tasks, supporting the claims from our theoretical analysis.
Specifically, on ALE, reward inflation achieved an $11.53\%$ improvement over the baseline, with Fed further increasing the gain to $12.59\%$.
On MuJoCo, based on the normalized score, reward inflation achieved an $18.51\%$ improvement over the non-inflation baseline, while the Fed variant further increased the improvement to $35.21\%$.
These results indicate more consistent improvements than related approaches, including reward normalization, prioritized experience replay, learning rate inflation, and recency-weighted sampling.
Additional experiments with TD3 and tabular Q-learning further examine the scope and recency-weighting effect of reward inflation, including its potential costs under deceptive rewards. Comparisons with plasticity methods further examine the relationship between dormant-neuron suppression and learning performance.

Our contributions are summarized as follows:
\begin{itemize}
    \item We propose reward inflation, a simple and nearly cost-free modification to RL training. We further introduce Fed, an adaptive variant that adjusts the inflation rate on the fly.
    \item We analyze how reward inflation induces recency weighting and sustains gradient signals, and investigate dormant-neuron suppression as a possible contributing mechanism for improved learning.
    \item We present empirical results demonstrating that reward inflation and Fed benefit a broad range of tasks across ALE and MuJoCo.
\end{itemize}

\section{Related works}
\paragraph{Reward shaping and normalization.}
These techniques have long been recognized as important factors in stabilizing and accelerating RL training.
A first line of work modifies the reward signal to ease optimization, including potential-based shaping~\citep{ng1999policy}, intrinsic reward shaping~\citep{yuan2023automatic, aubret2019survey}, meta-learned shaping~\citep{zou2019reward, hu2020learning}, and curriculum-based reward design~\citep{park2025sparse, freitag2024curriculum, anca2023achieving, shukla2023automaton}. 
A closely related family normalizes reward or value magnitudes for stable training, including reward normalization~\citep{henderson2018deep, schulman2017proximal}, return-based normalization~\citep{schaul2021return}, and value rescaling~\citep{van2016learning, hessel2019multi}, with extensions to temporally heterogeneous rewards~\citep{dann2021adapting}. 
Reward inflation also modifies the reward signal, but differs in two respects: it preserves reward semantics and operates along a temporal axis, gradually scaling reward magnitudes over training rather than reshaping them per-state.

\paragraph{Implicit weighting of replayed experience.}
Such weighting schemes have also been explored to improve sample efficiency and learning stability in off-policy RL.
Prioritized experience replay (PER) weights transitions by TD error~\citep{schaul2015prioritized, zha2019experience, zhang2020self}, with extensions based on state distribution~\citep{sun2020attentive}, freshness~\citep{ma2022fresher}, and learnability~\citep{sujit2023prioritizing}. 
Other approaches emphasize recent experience (ERE)~\citep{wang2019boosting}, learn sampling policies~\citep{zha2019experience}, decouple actor-critic sampling~\citep{saglam2023actor}, or prioritize via reward prediction errors~\citep{yamani2025reward}. 
Reward inflation similarly increases the influence of recent transitions on TD updates, but does so by progressively amplifying reward magnitudes rather than altering replay or sampling mechanisms.
Daley et al.~\citep{daley2024demystifying} analyze the benefits of recency-based temporal credit assignment in TD return estimators, providing a conceptual connection to our approach, which instead weights replayed transitions according to their collection time.

\paragraph{Plasticity in deep RL.}
Recent work has identified a range of pathologies in deep RL, including dormant neurons~\citep{sokar2023dormant}, capacity loss~\citep{lyle2023understanding}, and primacy bias~\citep{nikishin2022primacy}. These collectively reflect a loss of plasticity and pose a growing obstacle to sustained learning.
Proposed remedies include periodic resets of network parameters~\citep{nikishin2022primacy, dohare2024loss}, regularization toward initialization~\citep{lyle2024disentangling}, and architectural interventions. 
Reward inflation offers an indirect alternative: by gradually amplifying gradient magnitudes, it sustains the effective update signal, which can equivalently be interpreted as maintaining the effective learning rate~\citep{lyle2024normalization}, thereby mitigating the emergence of dormant neurons without explicit resetting or architectural change.

\section{Method}
\label{sec:Method}
In this section, we present the notation and formulation of reward inflation, followed by the introduction of Fed, an adaptive inflation control mechanism.

\subsection{Preliminaries}
We formulate the RL problem as a Markov decision process (MDP), defined by $\mathcal{M} = \{\mathcal{S}, \mathcal{A}, \mathcal{P}, \mathcal{R}, \gamma\}$, where $\mathcal{S}$ and $\mathcal{A}$ denote the state and action spaces, $\mathcal{P}$ the transition function, $\mathcal{R}$ the reward function, and $\gamma$ the discount factor.
A policy $\pi \in \Pi$ maps states to action distributions.
At each timestep $t$, the agent samples an action $a_t \sim \pi(\cdot \mid s_t)$, transitions to $s_{t+1} \sim \mathcal{P}(\cdot \mid s_t, a_t)$, and receives reward $r_t$.
The return is defined as $G_t = \sum_{k=t}^{T} \gamma^{k-t} r_k$, where $T$ is the episode horizon.
The goal of RL is to find the optimal policy $\pi^* = \arg\max_{\pi} \mathbb{E}_\pi[G_0]$.

\subsection{Reward Inflation}
We consider off-policy RL with experience replay. Reward inflation is applied before rewards are inserted into the replay buffer, and stored rewards remain unchanged thereafter. Under reward inflation, the agent learns from the nominal reward $r^\prime_t$ rather than the real reward $r_t$.

We denote \textit{global step} $g\in\mathbb{N}_0$ as a unique index that counts the total number of environment interactions throughout training.
The inflation is represented as \textit{inflation rate} $\rho_Y(g\,;\,Y)$, where $Y$ denotes the number of global steps to incrementally apply the inflation rate.
From the definition, \textit{per-step inflation rate} $\tilde{\rho}(g)$ can be derived as
\begin{gather}
    \tilde{\rho}(g) \coloneqq (1+\rho_Y(g\,;\,Y))^{\tfrac{1}{Y}}-1.
\end{gather}
For example, a constant inflation rate of $2\%$ over $Y=10^5$ steps corresponds to a per-step rate of $\tilde{\rho}=1.02^{1/10^5}-1\approx1.98\times10^{-7}$. Compounding this per-step rate over $10^5$ steps increases the reward scale by $2\%$.
This implies that, if the inflation rate remains constant, the reward scale will increase by $2\%$ after $10^5$ steps.
Then we define \textit{inflation level} $I(g)>0$ at global step $g$ as 
\begin{gather}
I(g) \coloneqq \textstyle\prod_{k=1}^{g} \left(1+\tilde{\rho}(k)\right), 
\qquad
r_t^\prime(g) \coloneqq r_t \cdot I(g),
\end{gather}
where $I(0)=1$ by convention.
At any step $g$, the inflation level $I(g)$ represents the cumulative reward scaling relative to the initial value $I(0)$, under which the nominal and real rewards are identical.

As training progresses, the inflation level increases exponentially under a constant positive per-step inflation rate $\tilde{\rho}>0$, a condition referred to as \textit{positive inflation}.
If $\tilde{\rho} = 0$, the inflation level remains constant, reducing the setting to standard RL when maintained throughout training.
For $-1 < \tilde{\rho} < 0$, the reward scale gradually contracts over time.
This case, termed \textit{negative inflation} (or deflation), produces effects opposite to those of positive inflation.
We exclude the degenerate case of $\tilde{\rho} \leq -1$, in which rewards vanish or invert, thereby distorting the original learning problem.

\subsection{Fed: Adaptive Inflation Control}
Building on reward inflation, we further propose \textit{Fed}, an adaptive heuristic that controls the inflation rate during training.
Inspired by the role of the Federal Reserve in stabilizing the economy by lowering interest rates during downturns and raising them during overheating, Fed adjusts the inflation rate in response to the recent dynamics of the gradient norm.

Fed monitors the gradient norm $\mathcal{G}(\phi) = \lVert\nabla_\phi J(\phi)\rVert_2$, where $J(\phi)$ is the actor loss for actor-critic methods or the Q-function loss for value-based methods.
We use an exponential moving average (EMA) over a target sequence $\{x_k\}$:
\begin{equation}
    m_k(x) = \beta \cdot m_{k-1}(x) + (1-\beta) \cdot x_k.
\end{equation}
Fed maintains short- and long-term EMAs $m^s_k$ and $m^l_k$ of $\mathcal{G}(\phi)$, with $\beta_s < \beta_l$ so that $m^s_k$ tracks recent fluctuations while $m^l_k$ captures longer-term trends.
Their ratio
\begin{equation}
    m^{s/l}_k = m^s_k / m^l_k
\end{equation}
serves as a state indicator: $m^{s/l}_k > 1$ signals overheating (gradients growing faster than the long-term trend), while $m^{s/l}_k < 1$ signals a downturn.

To control inflation, Fed adjusts $\rho_Y$ as
\begin{equation}
    \rho^{\text{new}}_Y = \rho^\ast_Y - \lambda_1 (m^{s/l}_k - 1) - \lambda_2 (m^{s/l}_k - m^{s/l}_{k-1}),
    \label{eq:new_inflation_rate}
\end{equation}
where $\rho^\ast_Y$ is the target inflation rate.
The first correction term, scaled by $\lambda_1$, is a proportional response to the deviation $m^{s/l}_k - 1$,
reducing $\rho_Y$ to slow the amplification when $m^{s/l}_k>1$, and vice versa.
The second term, scaled by $\lambda_2$, is a derivative response to the rate of change $m^{s/l}_k - m^{s/l}_{k-1}$, providing anticipatory damping against rapid swings.
This update rule is inspired by the Taylor Rule~\citep{taylor1993discretion} from monetary policy and extends it with a derivative term, yielding a PD-style controller that directly modulates the inflation rate.

To prevent abrupt changes, we apply a soft update for the next $\rho_Y$:
\begin{equation}
    \rho_Y = \tau_\rho \cdot \rho^{\text{old}}_Y + (1-\tau_\rho) \cdot \rho^{\text{new}}_Y,
\end{equation}
where $\rho^{\text{old}}_Y$ is the previous rate and $\tau_\rho \in [0, 1)$ is the soft update coefficient.
Every $Y$ global training steps, Fed performs $K$ updates of $m^{s/l}_k$ and $\rho_Y$.

\section{Theoretical Analysis of Reward Inflation} 
In this section, we analyze the several effects of reward inflation on RL under fixed inflation.

\paragraph{Optimality under uniform reward scaling.}
A natural concern is whether scaling rewards alters the problem being solved. If all rewards are multiplied by the same positive factor $I(g)$, the resulting reward function $r'(s,a)=I(g)r(s,a)$ induces the same optimal policy under the standard return objective, since positive scaling leaves the arg max over policies unchanged. During learning, however, the replay buffer contains rewards scaled at different collection times, and its overall reward scale evolves with training progress. We now examine how these dynamics reshape TD learning.

\paragraph{Recency-weighting on TD update.}
To investigate how the inflated reward $r^\prime = I(g)\cdot r$ behaves in $Q$-learning, we derive the TD-update equation at the gradient level.
Throughout this analysis, we treat $\theta$ as the parameters at the current training step (held fixed). Each transition in the replay buffer $\mathcal{D}$ takes the form $(s,a,r^\prime,s^\prime)$ and is associated with the global step $g$ at which it was collected, so that $r^\prime = I(g)\cdot r$ for the underlying raw reward $r$.

We further assume:
\begin{itemize}
    \item[(A1)] The replay buffer $\mathcal{D}$ induces an approximately stationary joint distribution over transitions and collect times, so that the marginal sample distribution and the inflation factor $I(g)$ are approximately decoupled. In particular, the conditional distribution of $(s,a,r,s^\prime)$ given collect time $g$ is approximately equal to its marginal under $\mathcal{D}$.
    \item[(A2)] The bootstrapped term $\max_{a^\prime} Q(s^\prime,a^\prime)$ is treated as a stop-gradient target (e.g., a target network), so $\nabla_\theta$ does not pass through it.
\end{itemize}

The TD-update for $Q_\theta$ is described by the loss
\begin{equation}
    L(\theta)=\mathbb{E}_{(s,a,r^\prime,s^\prime)\sim \mathcal{D}}\!\left[\tfrac{1}{2}\big(y-Q_\theta(s,a)\big)^2\right],
\end{equation}
where $y = r^\prime + \gamma \max_{a^\prime} Q(s^\prime,a^\prime)$ denotes the TD-target. Its gradient is
\begin{equation}
    \nabla_\theta L(\theta)=-\mathbb{E}_{(s,a,r^\prime,s^\prime)\sim \mathcal{D}}\!\left[\big(y - Q_\theta(s,a)\big)\nabla_\theta Q_\theta(s,a)\right].
    \label{eq:td_loss_derivative}
\end{equation}

The gradient update rule for $\theta$ is $\theta \leftarrow \theta - \eta\nabla_\theta L(\theta)$, where $\eta$ denotes the learning rate. To isolate the effect of collect-time inflation, we examine the expected update contribution from transitions collected at a specific global step $g$. Letting $\mathcal{D}\mid g$ denote the conditional distribution over $\mathcal{D}$ given collect time $g$, we define
\begin{equation}
    \Delta\theta(g) = \eta\, \mathbb{E}_{(s,a,r,s^\prime)\sim \mathcal{D}\mid g}\!\left[\big(I(g)\cdot r + \gamma\max_{a^\prime}Q(s^\prime,a^\prime) - Q_\theta(s,a)\big)\nabla_\theta Q_\theta(s,a)\right],
\end{equation}
and let $\lVert\Delta\theta(g)\rVert$ denote the corresponding update strength.

Under (A1), the conditional distribution $\mathcal{D}\mid g$ shares its marginal over $(s,a,r,s^\prime)$ with $\mathcal{D}$, so $\Delta\theta(g)$ decomposes as
\begin{align}
    \Delta\theta(g) 
    &= \underbrace{\eta\, I(g)\, \mathbb{E}_{(s,a,r,s^\prime)\sim \mathcal{D}}\!\left[r\, \nabla_\theta Q_\theta(s,a)\right]}_{\text{reward-driven component}} \notag \\
    &\quad + \underbrace{\eta\, \mathbb{E}_{(s,a,r,s^\prime)\sim \mathcal{D}}\!\left[\big(\gamma\max_{a^\prime}Q(s^\prime,a^\prime) - Q_\theta(s,a)\big)\, \nabla_\theta Q_\theta(s,a)\right]}_{\text{bootstrapping component}}.
\end{align}

With $Q_\theta$ held fixed, the reward-driven component explicitly depends on the collection-time inflation factor $I(g)$, while the expected bootstrapping component has no explicit dependence on collection time $g$ under (A1).
Consequently, when comparing two collection times $g_1<g_2$, the bootstrapping components cancel and only the reward-driven components contribute to the difference:
\begin{align}
    \Delta\theta(g_2)-\Delta\theta(g_1) 
    &= \big(I(g_2)-I(g_1)\big)\cdot \eta\, \mathbb{E}_{(s,a,r,s^\prime)\sim \mathcal{D}}\!\left[r\, \nabla_\theta Q_\theta(s,a)\right].
\end{align}
Taking norms,
\begin{equation}
    \big\lVert \Delta\theta(g_2)-\Delta\theta(g_1) \big\rVert 
    = \big|I(g_2)-I(g_1)\big| \cdot \eta\, \big\lVert \mathbb{E}_{(s,a,r,s^\prime)\sim \mathcal{D}}\!\left[r\, \nabla_\theta Q_\theta(s,a)\right] \big\rVert.
    \label{eq:update_gap}
\end{equation}

By \eqref{eq:update_gap}, the update discrepancy scales linearly with $|I(g_2)-I(g_1)|$, so the transition with the larger inflation factor contributes the larger reward-driven component.
Under \textit{positive inflation}, $I(g)$ increases monotonically with $g$, so recently collected transitions ($g_2$) tend to dominate older ones ($g_1$).
Conversely, under \textit{negative inflation} (deflation), older transitions tend to dominate.

\paragraph{Time-invariance of inflation weighting.}
For a fixed collection-time difference $g_2-g_1$, the absolute inflation gap $|I(g_2)-I(g_1)|$ grows under constant positive inflation but shrinks under constant deflation as training progresses.
The \emph{shape} of recency-weighting in the reward-driven update, however, is invariant to absolute training time: isolating the reward-driven component $\Delta\theta_{\text{rew}}(g) := \eta\, I(g)\, \mathbb{E}_{\mathcal{D}}[r\,\nabla_\theta Q_\theta(s,a)]$ from the decomposition of $\Delta\theta(g)$, its ratio between any two transitions reduces to
\begin{equation}
    \frac{\Delta\theta_{\text{rew}}(g_2)}{\Delta\theta_{\text{rew}}(g_1)} = \frac{I(g_2)}{I(g_1)} = (1+\tilde{\rho})^{\,g_2-g_1}.
\end{equation}
Under exponential inflation $I(g)=(1+\tilde{\rho})^g$ with $\tilde{\rho}>-1$, this ratio depends only on the relative time difference $g_2-g_1$, not on the absolute training step—a consequence of the multiplicative closure $I(g+\Delta)=(1+\tilde{\rho})^\Delta\, I(g)$.
The shape of recency-weighting in the reward-driven update therefore remains stationary across training: two transitions $\Delta$ steps apart always carry the same relative weight $(1+\tilde{\rho})^\Delta$, with positive inflation ($\tilde{\rho}>0$) weighting recent transitions more heavily and deflation ($-1<\tilde{\rho}<0$) doing the reverse.
In contrast, linear or polynomial inflation profiles induce a recency-weighting whose shape flattens over time, since $I(g_2)/I(g_1)\to 1$ as $g_1,g_2\to\infty$ for fixed $g_2-g_1$.

\paragraph{Cumulative effect via bootstrapping.}
While the above analysis isolates the instantaneous effect of inflation on a single update step, the bootstrapping component absorbs inflation indirectly over the course of training. Since $Q_\theta$ itself is learned from inflated rewards stored in the buffer, the term $\gamma \max_{a^\prime} Q_\theta(s^\prime,a^\prime)$ progressively reflects the average inflation level
\begin{equation}
    \bar{I} := \mathbb{E}_{(s,a,r,s^\prime)\sim \mathcal{D}}[I(g)]
\end{equation}
of the buffer. Under (A1), in a quasi-stationary regime where $Q_\theta$ approximately tracks the fixed point of the inflated Bellman operator, unrolling the Bellman recursion yields the approximate scaling
\begin{equation}
    Q_\theta(s,a) \;\approx\; \bar{I}\cdot Q^*(s,a),
\end{equation}
where $Q^*$ denotes the optimal $Q$-function under the un-inflated reward $r$. With a finite (e.g., FIFO) replay buffer, $\bar{I}$ tracks the recent inflation level rather than the full training history. Consequently, $\bar{I}$ drifts as training progresses—upward under positive inflation and downward under deflation—inducing a non-stationary shift in the magnitude of $Q_\theta$ in the corresponding direction. In this sense, the reward-driven component injects inflation locally at each transition, while the bootstrapping component cumulatively scales the entire $Q$-function by the buffer-averaged inflation $\bar{I}$.

\paragraph{Gradient norm scaling under inflation.}
We now examine how the gradient norm $\lVert\nabla_\theta L(\theta)\rVert$ scales with the buffer-averaged inflation $\bar{I}$. Under the quasi-stationary scaling $Q_\theta\approx\bar{I}\cdot Q^*$ established earlier, the parametric gradient inherits the same factor, $\nabla_\theta Q_\theta \approx \bar{I}\cdot \nabla_\theta Q^*$.
Since the TD-error $y - Q_\theta(s,a)$ is itself of order $\bar{I}$ (with $I(g)\approx\bar{I}$ on average within $\mathcal{D}$), the loss gradient scales as
\begin{equation}
    \nabla_\theta L(\theta) = -\mathbb{E}_{\mathcal{D}}\!\left[\big(y - Q_\theta(s,a)\big)\, \nabla_\theta Q_\theta(s,a)\right] 
    \;\Longrightarrow\; \lVert \nabla_\theta L(\theta)\rVert \;\propto\; \bar{I}^{\,2}.
\end{equation}

A similar analysis for the policy gradient yields a different scaling. For the advantage-based estimator $\nabla_\phi J(\phi)=\mathbb{E}[\nabla_\phi\log\pi_\phi(a|s)\cdot A^\pi(s,a)]$, the score function $\nabla_\phi\log\pi_\phi$ is independent of the reward scale, while the advantage $A^\pi=Q^\pi-V^\pi$ inherits the $\bar{I}$ scaling of the value functions. Hence
\begin{equation}
    \lVert\nabla_\phi J(\phi)\rVert \;\propto\; \bar{I}.
\end{equation}
In both cases, gradient norms grow with $\bar{I}$ under \textit{positive inflation}, and shrink under \textit{negative inflation} (deflation).

\paragraph{Convergence considerations.}
The approximate scaling in Eq.~(15) describes tracking a moving value scale, rather than convergence to a fixed Q-function. Under persistent positive inflation, the buffer-averaged reward scale can grow without bound, so convergence to a finite Q-function is generally not guaranteed. If the reward scale instead approaches a finite positive limit and the replay buffer refreshes, the learning problem approaches a stationary, uniformly scaled MDP. Further details are provided in Appendix~\ref{app:convergence}.

\paragraph{Connection to plasticity loss.}
The cumulative gradient-norm scalings $\lVert\nabla_\theta L\rVert\propto\bar{I}^{\,2}$ and $\lVert\nabla_\phi J(\phi)\rVert \propto \bar{I}$ derived above suggest a mechanistic connection to plasticity loss in deep RL, where the fraction of dormant neurons grows over the course of training~\citep{lyle2023understanding,sokar2023dormant}.
By gradually amplifying gradient magnitudes, reward inflation sustains the effective learning rate~\citep{lyle2024normalization} that would otherwise diminish as the policy saturates.
This is consistent with prior findings that weak gradients are a primary cause of dormant neurons~\citep{arnekvist2020effect,horuz2025resurrection}. 

\section{Experiment}
We organize our experiments around the following questions:
(1) Do reward inflation and Fed improve learning performance?
(2) How does reward inflation compare with related approaches?
(3) What evidence supports the proposed mechanisms, and what trade-offs accompany reward inflation?

\subsection{Experimental Setup}
\label{sec:ExperimentalSetup}
We conduct experiments on 30 games from the Arcade Learning Environment (ALE) 2600~\citep{bellemare2013arcade} using DQN~\citep{mnih2015human} to assess the overall performance impact of reward inflation.
We further evaluate on eight MuJoCo tasks~\citep{todorov2012mujoco} (six locomotion and two manipulation) using Soft Actor-Critic (SAC)~\citep{haarnoja2018soft} to corroborate our theoretical analysis and conduct ablation studies, and using TD3~\citep{dankwa2019twin} as an alternative RL algorithm.
Additionally, we use tabular Q-learning~\citep{watkins1992q} with a replay buffer to examine the recency-weighting effect of reward inflation in a more controlled setting without neural function approximation.

Results are averaged over multiple random seeds, with counts specified alongside the corresponding results.
All agents are evaluated using the original environment rewards without inflation.
The $\pm$ values in the tables denote standard deviations across seeds.
The main ALE and MuJoCo experiments run for 10M and 3M environment steps, respectively.
For the inflation rate $\rho_Y(g~;~Y)$, we set $Y=10^5$ steps.
We set Fed's inflation target at $2\%$.
Detailed RL hyperparameters, Fed hyperparameters, and comparison-baseline implementations are provided in Appendices~\ref{app:rl_hyperparameters}, \ref{app:fed_hyperparameters}, and~\ref{app:comparison_baselines}, respectively.

\subsection{Performance Evaluation}
On both ALE and MuJoCo, we first compare baseline agents against two variants: one trained with a fixed $2\%$ inflation, and one trained with Fed at a $2\%$ target inflation rate.
We then broaden the comparison on MuJoCo to additional inflation rates and to related approaches.

\subsubsection{Arcade Learning Environment}

\begin{table}[t]
    \caption{Performance summary on 30 ALE tasks. Tied methods each receive a win.}
    \centering
    \small
    \begin{tabular}{lrrr}
        \toprule
        \textbf{Metric} & \textbf{0\%} & \textbf{2\%} & \textbf{Fed (2\%)} \\
        \midrule
        \# Winning Tasks & 2 & \textbf{16} & 13 \\
        Mean HNS & 219.5 & 244.8 & \textbf{247.2} \\
        \bottomrule
    \end{tabular}
    \label{tab:atari_inflation_summary}
\end{table}

\begin{table}[t]
\centering
\caption{Performance comparison under varying inflation rates (upper rows) and other methods (lower rows) across MuJoCo tasks. Bold and underlined means indicate the best and second-best results across all methods, respectively. The dagger ($\dagger$) marks the best value among fixed inflation rates. The last column reports the average within-table min-max normalized score across tasks. Returns are evaluated without reward inflation.}
\small
\setlength{\tabcolsep}{0pt}
\begin{tabular*}{\linewidth}{@{\extracolsep{\fill}} l c c c c c c c c c @{}}
\toprule
\textbf{Method} & \textbf{HalfCheetah} & \textbf{Walker2d} & \textbf{Humanoid} & \textbf{Swimmer} & \textbf{Hopper} & \textbf{Ant} & \textbf{Pusher} & \textbf{Reacher} & \textbf{NS} \\
\midrule
-2\%
& 14020{\scriptsize $\pm$1082}
& 4342{\scriptsize $\pm$461}
& 5300{\scriptsize $\pm$176}
& 92{\scriptsize $\pm$43}
& 2421{\scriptsize $\pm$460}
& 3929{\scriptsize $\pm$1090}
& -24.25{\scriptsize $\pm$4.10}
& -3.46{\scriptsize $\pm$0.23}
& 36.8 \\

0\%
& 15241{\scriptsize $\pm$734}
& 4399{\scriptsize $\pm$527}
& 5826{\scriptsize $\pm$313}
& 116{\scriptsize $\pm$36}
& 2869{\scriptsize $\pm$299}
& 4763{\scriptsize $\pm$1642}
& -23.28{\scriptsize $\pm$2.89}
& \textbf{-3.41}{\scriptsize $\pm$0.23}$^\dagger$
& 68.6 \\

1\%
& \textbf{15707}{\scriptsize $\pm$715}$^\dagger$
& 5007{\scriptsize $\pm$560}
& 5859{\scriptsize $\pm$363}
& 121{\scriptsize $\pm$25}
& 2821{\scriptsize $\pm$406}
& 5245{\scriptsize $\pm$1312}$^\dagger$
& -22.71{\scriptsize $\pm$2.87}
& \underline{-3.42}{\scriptsize $\pm$0.23}
& 79.8 \\

2\%
& 15607{\scriptsize $\pm$372}
& \underline{5102}{\scriptsize $\pm$520}$^\dagger$
& 5861{\scriptsize $\pm$199}
& \underline{136}{\scriptsize $\pm$11}$^\dagger$
& 3040{\scriptsize $\pm$499}
& 4346{\scriptsize $\pm$1267}
& \underline{-22.07}{\scriptsize $\pm$1.58}$^\dagger$
& \underline{-3.42}{\scriptsize $\pm$0.24}
& \underline{81.3}$^\dagger$ \\

3\%
& 15091{\scriptsize $\pm$927}
& 4690{\scriptsize $\pm$624}
& \underline{5873}{\scriptsize $\pm$306}$^\dagger$
& 126{\scriptsize $\pm$11}
& 3035{\scriptsize $\pm$445}
& 4392{\scriptsize $\pm$1298}
& -22.58{\scriptsize $\pm$1.58}
& -3.44{\scriptsize $\pm$0.24}
& 71.5 \\

5\%
& 14484{\scriptsize $\pm$996}
& 4615{\scriptsize $\pm$518}
& 5545{\scriptsize $\pm$492}
& 131{\scriptsize $\pm$12}
& \underline{3161}{\scriptsize $\pm$346}
& 4795{\scriptsize $\pm$1089}
& -22.20{\scriptsize $\pm$0.89}
& -3.45{\scriptsize $\pm$0.23}
& 71.5 \\

10\%
& 14186{\scriptsize $\pm$1200}
& 4167{\scriptsize $\pm$650}
& 5209{\scriptsize $\pm$1241}
& 129{\scriptsize $\pm$19}
& \textbf{3166}{\scriptsize $\pm$288}$^\dagger$
& 4248{\scriptsize $\pm$1219}
& -24.60{\scriptsize $\pm$1.47}
& -3.54{\scriptsize $\pm$0.33}
& 46.6 \\

Fed (2\%)
& \underline{15621}{\scriptsize $\pm$991}
& \textbf{5395}{\scriptsize $\pm$452}
& \textbf{6048}{\scriptsize $\pm$436}
& \textbf{143}{\scriptsize $\pm$13}
& 3105{\scriptsize $\pm$293}
& 5180{\scriptsize $\pm$1352}
& \textbf{-21.99}{\scriptsize $\pm$1.23}
& -3.43{\scriptsize $\pm$0.24}
& \textbf{92.8} \\
\midrule
2\% (lr)
& 14566{\scriptsize $\pm$1006}
& 4922{\scriptsize $\pm$515}
& 5417{\scriptsize $\pm$298}
& 77{\scriptsize $\pm$41}
& 2899{\scriptsize $\pm$359}
& 4458{\scriptsize $\pm$1888}
& -23.95{\scriptsize $\pm$2.88}
& \underline{-3.42}{\scriptsize $\pm$0.24}
& 59.2 \\

PER
& 14734{\scriptsize $\pm$1934}
& 4797{\scriptsize $\pm$592}
& 4542{\scriptsize $\pm$2290}
& 57{\scriptsize $\pm$23}
& 2683{\scriptsize $\pm$529}
& \textbf{5762}{\scriptsize $\pm$979}
& -26.41{\scriptsize $\pm$2.98}
& -3.43{\scriptsize $\pm$0.24}
& 48.7 \\

RN
& 15022{\scriptsize $\pm$436}
& 5036{\scriptsize $\pm$466}
& 5578{\scriptsize $\pm$285}
& 118{\scriptsize $\pm$13}
& 3149{\scriptsize $\pm$272}
& 5017{\scriptsize $\pm$1666}
& -29.17{\scriptsize $\pm$13.37}
& -3.46{\scriptsize $\pm$0.24}
& 64.1 \\

RTN
& 11877{\scriptsize $\pm$369}
& 4454{\scriptsize $\pm$571}
& 4451{\scriptsize $\pm$633}
& 87{\scriptsize $\pm$21}
& 2742{\scriptsize $\pm$463}
& 4046{\scriptsize $\pm$1603}
& -23.41{\scriptsize $\pm$2.05}
& -3.45{\scriptsize $\pm$0.13}
& 32.1 \\

RWS
& 15482{\scriptsize $\pm$687}
& 5000{\scriptsize $\pm$546}
& 5730{\scriptsize $\pm$362}
& 104{\scriptsize $\pm$40}
& 2923{\scriptsize $\pm$469}
& \underline{5466}{\scriptsize $\pm$1318}
& -23.32{\scriptsize $\pm$3.55}
& -3.43{\scriptsize $\pm$0.20}
& 76.8 \\
\bottomrule
\end{tabular*}
\label{tab:inflation_performance}
\end{table}

To investigate how reward inflation and Fed generally affect training performance, we evaluated 30 ALE Atari 2600 games using standard DQN.
Table~\ref{tab:atari_inflation_summary} compares the DQN baseline, $2\%$ reward inflation, and Fed with a target inflation rate of $2\%$, summarizing the average performance over the final 1M training steps, with the last row reporting the average Human-Normalized Score (HNS)~\citep{badia2020agent57}.
Results are averaged over 8 random seeds per condition.
Surprisingly, despite the small modification, reward inflation significantly improved performance in most games, yielding an average HNS improvement of $11.53\%$ over the baseline (mean baseline-normalized improvement: $17.28\%$, 95\% CI $[5.16\%, 29.41\%]$; one-sample $t$-test against zero, $t(29) = 2.92$, $p = 0.0068$).
Fed further improved performance, achieving a $12.59\%$ HNS gain over the baseline (mean baseline-normalized improvement: $19.60\%$, 95\% CI $[2.35\%, 36.85\%]$; $t(29) = 2.32$, $p = 0.027$).
These results highlight the beneficial impact of reward inflation.
The full table, learning curves, and bootstrap CIs are provided in Table~\ref{tab:atari_inflation_full}, Figure~\ref{fig:atari_learning_curves}, and Figure~\ref{fig:bootstrapCI_taskwise_atari}, respectively.

\subsubsection{MuJoCo}
We further evaluated on eight MuJoCo tasks.
This time, we evaluated various inflation rates to examine how each rate differently affects training, using rates of $-2\%$, $0\%$, $1\%$, $2\%$, $3\%$, $5\%$, and $10\%$.
For comparison with related approaches, we additionally evaluated $2\%$ learning rate (LR) inflation, PER, reward normalization (RN), return normalization (RTN), and recency-weighted sampling (RWS).
The $2\%$ (lr) baseline increases learning rates rather than scaling rewards.
RWS, inspired by ERE, is designed to match the recency-weighting effect of $2\%$ reward inflation (see Appendix~\ref{app:comparison_baselines}).

Table~\ref{tab:inflation_performance} reports the episode return averaged over the final $300$K training steps.
Results are averaged over 10 random seeds per condition.
The last column reports the average within-table min-max normalized score (NS) across tasks, where normalization is performed independently for each task using the minimum and maximum scores among all compared methods.
On MuJoCo, compared to the baseline ($0\%$), $2\%$ inflation improved the NS by $18.51\%$, while Fed with a target inflation rate of $2\%$ further increased the improvement to $35.21\%$.
This result again highlights the general benefits of reward inflation.

\paragraph{Performance under variety of inflation rates.}
Overall, based on NS, while $1$--$5\%$ inflation showed improvements over the baseline, the two extremes, $-2\%$ deflation and $10\%$ inflation, resulted in severe degradation ($-46.43\%$ and $-32.16\%$, respectively).
Since $-2\%$ deflation induces reversed recency weighting, emphasizing older samples and shrinking gradient norms, performance was generally degraded.
Conversely, the degradation under $10\%$ inflation suggests that while moderate inflation can act as a healthy stimulus, excessive inflation introduces side effects that hurt training.
However, since $10\%$ inflation achieved the best performance on \texttt{Hopper}, where scores increased alongside the inflation rate, the most beneficial rate appears to be task-dependent.
Statistically, the set of per-task best rates yielded a mean baseline-normalized improvement of $5.40\%$ (95\% CI $[0.30\%, 10.50\%]$; one-sample $t$-test against zero, $t(7) = 2.50$, $p = 0.041$), while Fed yielded $7.31\%$ ($[0.83\%, 13.79\%]$; $t(7) = 2.67$, $p = 0.032$).
Bootstrap CIs are also provided in Figure~\ref{fig:bootstrapCI_taskwise_mujoco}.

\paragraph{Comparison with related approaches.}
We compare reward inflation against several methods that share a related mechanism but intervene at different stages or through different signals.
While reward inflation induces an implicit recency weighting that assigns a different effective learning rate to each sample depending on its recency, LR inflation raises the update magnitude uniformly across all samples, degrading performance well below the baseline.
PER likewise reshapes the effective gradient signal, but does so by selectively up-weighting transitions with large TD errors; while this yields the strongest result on \texttt{Ant}, the gains are inconsistent across tasks.
RN and RTN also rescale rewards, but use running statistics rather than an explicit inflation schedule. Their performance varies across tasks, with neither method exceeding the baseline's aggregate NS.
While RWS yields only marginal gains over the baseline, moderate reward inflation and Fed deliver larger and more consistent improvements across tasks.

\paragraph{Results with TD3.}
To examine whether the benefits of reward inflation extend beyond SAC, we additionally evaluated TD3 with and without $2\%$ inflation on the same MuJoCo tasks over 8 random seeds per condition (Table~\ref{tab:td3_results}).
Inflation yielded modest AUC gains on seven of the eight tasks, while improvements in final scores were smaller and less consistent. These results suggest that the benefit can also occur without entropy-temperature tuning, although the larger gains observed with SAC may partly reflect its interaction with reward scaling.
\begin{figure*}
  \centering
  \includegraphics[width=1.0\linewidth]{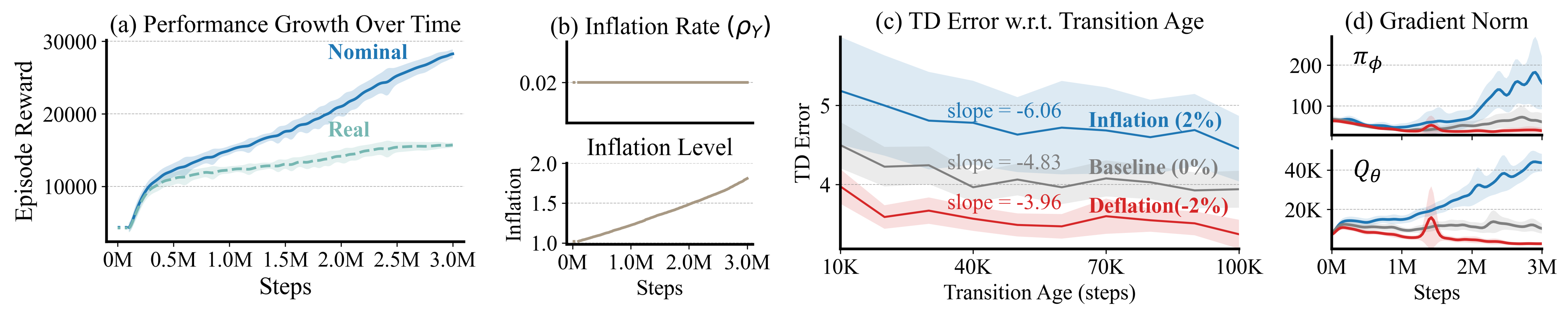}
  \caption{(a) The nominal and real performance growth over training time under $2\%$ inflation. (b) The inflation rate and inflation level $I(g)$ over the training time. (c) The TD error w.r.t. the transition age in replay buffer, across the inflation rates. (d) The changes of gradient norms during training.}
  \label{fig:characteristics}
\end{figure*}

\subsection{Empirical Analysis of Reward Inflation}

\paragraph{Nominal and real episode reward.}
Maximizing nominal return is one distinctive property of reward inflation.
Figure~\ref{fig:characteristics}(a,b) illustrates the gap between nominal and real episode rewards, alongside the change in the inflation level $I(g)$ during training.
This example uses \texttt{HalfCheetah-v4} with a $2\%$ inflation rate.
The nominal episode reward increased consistently faster than the real episode reward, in proportion to $I(g)$.
Note that the real episode reward is reported as the performance metric.
\begin{figure}[t]
  \centering
  \begin{minipage}{0.48\columnwidth}
    \centering
    \includegraphics[width=\linewidth]{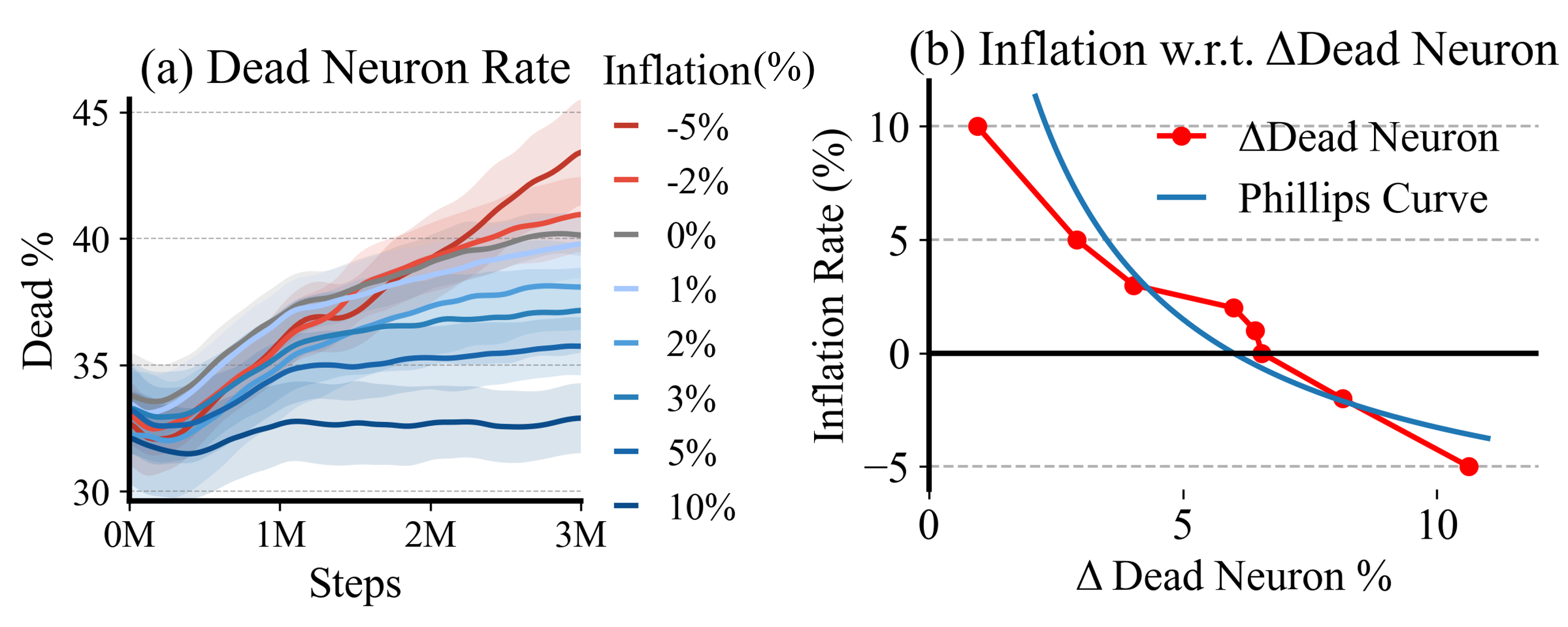}
    \caption{(a) The changes in dormant (dead) neuron rate during training from deflation (-$5\%$) to inflation ($10\%$). (b) The total increased rate of dead neurons under each inflation rate.}
    \label{fig:dead_neurons}
  \end{minipage}
  \hfill 
  \begin{minipage}{0.48\columnwidth}
    \centering
    \includegraphics[width=\linewidth]{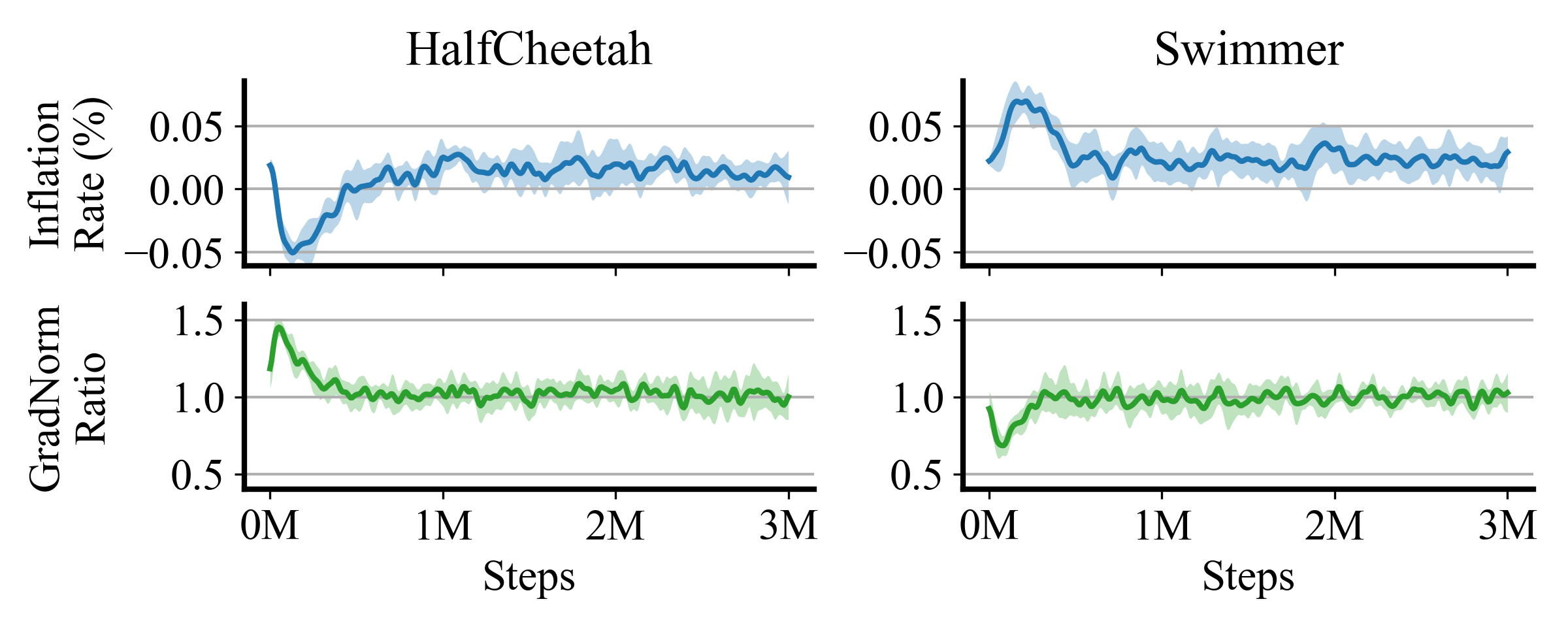}
    \caption{Inflation rates (top) adjusted by Fed in response to the observed short-long gradient norm ratios $m^{s/l}$ (bottom), on \texttt{HalfCheetah-v4} and \texttt{Swimmer-v4}.}
    \label{fig:federl}
  \end{minipage}
\end{figure}

\paragraph{Recency weighting and gradient norms.}
To investigate the impact of reward inflation on recency weighting and gradient norms, we evaluated agents with $2\%$ (inflation), $0\%$ (baseline), and $-2\%$ (deflation) inflation rates on \texttt{HalfCheetah-v4}.
Recency weighting was quantified by analyzing the dependence of TD error on the age of transitions.
As shown in Figure~\ref{fig:characteristics}(c), older transitions, which have been sampled more frequently, tended to exhibit lower TD errors.
However, inflation amplified the TD error gap between older and more recent transitions, whereas deflation reduced it.
Figure~\ref{fig:characteristics}(d) further illustrates how the progression of gradient norms over time differs across the three inflation rates.
These observations support the findings from the theoretical analysis.

\paragraph{Impacts on dormant neurons.}
A dormant (or dead) neuron~\citep{he2015delving} is defined as a neuron that persistently outputs zero, contributing little or nothing to further learning.
To investigate how reward inflation affects dormant neurons, we tracked the proportion of dormant neurons in both the $Q_\theta$ and $\pi^\phi$ networks throughout training.

Figure~\ref{fig:dead_neurons}(a) shows how the proportion of dormant neurons evolves under inflation rates ranging from $-5\%$ to $10\%$ on \texttt{HalfCheetah-v4}.
Surprisingly, a clear negative correlation emerges: the final proportion of dormant neurons decreases as the inflation rate increases, with deflation yielding a higher proportion than the baseline.
Although dormant neurons are commonly associated with ReLU, similar behavior has been reported under other activations~\citep{gustineli2022survey}; the same pattern accordingly holds across other tasks and activation functions (Figures~\ref{fig:dead_relu_additional},~\ref{fig:dead_leakyrelu}, and~\ref{fig:dead_silu}).
Jointly observing Figure~\ref{fig:dead_neurons}(a) and Figure~\ref{fig:characteristics}(d) shows that under the baseline ($0\%$), the gradient norm generally decreases while the proportion of dormant neurons concurrently increases. The decrease in gradient norm is expected as the agent's predictions align more closely with its learning targets.
Positive inflation instead increases the gradient norm over training, suppressing the emergence of dormant neurons, whereas deflation accelerates gradient decay and produces a more rapid rise in dormant neurons.

Intuitively, the decrease in dormant neuron rates with increasing inflation in Figure~\ref{fig:dead_neurons}(b) resembles the Phillips curve~\citep{phillips1958relation}, which captures the inflation-unemployment trade-off and echoes Keynesian ideas of activating idle resources~\citep{samuelson1960analytical}. Analogously, reducing dormant neurons may improve network utilization.
However, just as hyperinflation is undesirable, the lowest dormant neuron rate does not imply the best performance: $10\%$ inflation yields the fewest dormant neurons but the worst overall result.
This may reflect a trade-off between network utilization~\citep{sokar2023dormant} and the benefits of sparsity~\citep{dufort2024maxwell}, although the present results do not isolate these factors.

We additionally compare reward inflation with related plasticity baselines ReDo~\citep{sokar2023dormant}, Reset~\citep{nikishin2022primacy}, and Shrink \& Perturb~\citep{ash2020warm} (Tables~\ref{tab:plasticity_baselines} and~\ref{tab:plasticity_dormancy}; implementation details in Appendix~\ref{app:comparison_baselines}).
These methods generally reduce dormant neurons more strongly, but their performance gains are not consistent across tasks.
This suggests that dormant-neuron reduction alone is insufficient to explain the performance benefits of reward inflation.
A HalfCheetah task-switch experiment also suggests a potential benefit for adaptation to changing objectives (Appendix~\ref{app:forward_backward}).

\paragraph{Fed dynamics.}
Fed adjusts the inflation rate in response to changes in the short-long gradient ratio $m^{s/l}$.
Figure~\ref{fig:federl} illustrates how Fed controls inflation in two contrasting cases.
In \texttt{HalfCheetah-v4}, $m^{s/l}$ surged rapidly at the beginning of training, indicating overheating caused by strong learning signals.
As this task is known for rapid early performance gains, Fed responded by sharply lowering the inflation rate, even temporarily entering deflation.
Once $m^{s/l}$ stabilized, the inflation rate recovered to the target level of $2\%$ and remained relatively stable thereafter.
In contrast, \texttt{Swimmer-v4} exhibited the opposite trend.
Due to its slow and reward-scarce early phase, $m^{s/l}$ initially declined, prompting Fed to increase the inflation rate to stimulate learning.
As training progressed and $m^{s/l}$ stabilized, the inflation rate gradually converged to the target level.

\subsection{Tabular Q-Learning: Recency Weighting and Its Trade-Offs}

\paragraph{Recency weighting and learning performance.}
To examine reward inflation without function approximation, we trained tabular Q-learning agents on slippery $8\times8$ FrozenLake for 2M steps over 32 seeds, using inflation rates from $0\%$ to $5{,}000\%$.
Algorithmic and environment details are provided in Appendix~\ref{app:tabular_settings}.
We measure recency weighting through the slope of mean log absolute TD error against transition age for reward-bearing transitions; a more negative slope indicates larger TD errors for recent transitions.

\begin{table}[t]
    \caption{TD-error recency and learning performance in tabular Q-learning on slippery $8\times 8$ FrozenLake. The TD-error slope is fitted to mean $\log|\delta|$ as a function of transition age; more negative values indicate stronger recency weighting. Success rate and learning-curve AUC are reported as mean $\pm$ standard deviation over 32 seeds.}
    \centering
    \small
    \begin{tabular}{rccc}
        \toprule
        \textbf{Inflation} & \textbf{TD-error slope} & \textbf{Success rate} & \textbf{AUC} \\
        \midrule
        0\%    & $-2.039\times 10^{-4}$ & $0.478 \pm 0.132$ & $0.407 \pm 0.014$ \\
        1\%    & $-7.645\times 10^{-4}$ & $0.483 \pm 0.166$ & $0.408 \pm 0.015$ \\
        5\%    & $-3.220\times 10^{-3}$ & $0.520 \pm 0.109$ & $0.422 \pm 0.015$ \\
        10\%   & $-6.464\times 10^{-3}$ & $0.507 \pm 0.109$ & $0.431 \pm 0.018$ \\
        50\%   & $-2.600\times 10^{-2}$ & $0.584 \pm 0.047$ & $0.475 \pm 0.015$ \\
        100\%  & $-3.753\times 10^{-2}$ & $0.585 \pm 0.030$ & $0.482 \pm 0.015$ \\
        500\%  & $-8.895\times 10^{-2}$ & $0.586 \pm 0.033$ & $\mathbf{0.484} \pm 0.014$ \\
        1000\% & $-1.231\times 10^{-1}$ & $\mathbf{0.594} \pm 0.031$ & $\mathbf{0.484} \pm 0.018$ \\
        5000\% & $-1.930 \times 10^{-1}$ & $0.574 \pm 0.035$ & $0.474 \pm 0.019$ \\
        \bottomrule
    \end{tabular}
    \label{tab:tabular_q_recency}
\end{table}

In Table~\ref{tab:tabular_q_recency}, the slope becomes more negative with inflation despite uniform replay sampling.
Success rate and AUC generally improve, although gains diminish at the highest rates.
These results support stronger recency weighting and faster learning in this tabular setting.

\paragraph{Trap-entry evaluation.}
However, stronger recency weighting may increase susceptibility to deceptive rewards.
We therefore introduce an easily reached low-reward goal alongside a harder-to-reach high-reward goal and compare two exploration schedules.
As shown in Table~\ref{tab:tabular_q_trap}, reward inflation increases trapping at high rates, while slower $\epsilon$ decay reduces it.
This suggests that inflation may require a larger exploration budget to avoid deceptive rewards.

\subsection{Limitations}
\label{subsect:limitations}
Several limitations remain.
First, our experiments cover finite training horizons.
Prolonged inflation can destabilize optimization through reward-scale growth, motivating stabilization or saturation mechanisms (see Appendix~\ref{app:convergence}).
Second, our current formulation relies on off-policy learning with replay, where transitions collected at different reward scales coexist in the buffer. Extending this mechanism to on-policy learning remains future work.
Third, the optimal inflation behavior appears to depend on task characteristics and training dynamics, suggesting that a universally optimal inflation schedule may not exist.
Finally, Fed currently relies on multiple hyperparameters and a heuristic control rule, motivating future work on more principled and adaptive inflation-control strategies.

\section{Conclusion}
In this paper, we introduced reward inflation, a simple temporal reward-scaling scheme that can act as a healthy stimulus for off-policy RL with experience replay. Experiments on ALE and MuJoCo demonstrate performance benefits across a broad range of tasks under moderate inflation, with Fed often improving upon fixed inflation. Our analysis characterizes implicit recency weighting and gradient scaling, while experiments identify dormant-neuron suppression as a possible contributing mechanism. Tabular Q-learning experiments further support the recency effect while revealing increased susceptibility to deceptive rewards at high inflation rates. These findings motivate further investigation into temporal reward dynamics as a lightweight approach to improving RL.

\section*{Acknowledgment}
This work was supported in part by the Korea Research Institute for Defense Technology Planning and Advancement (KRIT) grant funded by the Defense Acquisition Program Administration (DAPA) (No. KRIT-CT-23-003/10\%, Development of AI Researchers Based on Deep Reinforcement Learning and Establishment of Virtual Combat Experiment Environment), and in part by the IITP (RS-2021-II211343-GSAI/10\%, RS-2022-II220951-LBA/10\%, RS-2022-II220953-PICA/10\%, RS-2026-25553157-MIACC/10\%), the NRF (RS-2024-00353991-SPARC/10\%, RS-2023-00274280-HEI/10\%, RS-2024-00358416-AutoRL/10\%), the KEIT (RS-2025-25453780/10\%), and the KIAT (RS-2025-25460896/10\%), funded by the Korean government.

\bibliographystyle{unsrt} 
\bibliography{citations}

\newpage
\appendix

\section{Experimental Details}
\label{app:exp_details}
\subsection{RL Hyperparameters}
\label{app:rl_hyperparameters}

\paragraph{SAC.}
Unless otherwise specified, SAC experiments use the hyperparameters listed in Table~\ref{tab:hyperparameter}.
These settings largely follow the CleanRL implementation~\citep{huang2022cleanrl}.
We train for 3M environment steps to examine the effects of reward inflation over an extended training horizon.
We increase the warm-up period from 5K to 100K steps because the shorter warm-up occasionally produced poor performance for both the baseline and inflated agents.
For \textit{Hopper-v4} and \textit{Ant-v4}, we adjust the discount factor to reduce performance volatility.
The SAC baseline agents with these adjusted discount factors also outperform those using the default $\gamma=0.99$.
The task-specific discount factors in Table~\ref{tab:gamma} are used for both SAC and TD3.

\begin{table}[h!]
    \centering
    \caption{SAC hyperparameter configuration.}
    \label{tab:hyperparameter}
    \begin{tabular}{ll}
        \toprule
        \textbf{Parameter} & \textbf{Value} \\
        \midrule
        Total environment steps & 3M \\
        Replay buffer size & 1M \\
        Target update coefficient ($\tau$) & 0.005 \\
        Batch size & 256 \\
        Warm-up steps & 100K \\
        Learning rate ($Q_\theta$) & 0.001 \\
        Learning rate ($\pi_\phi$) & 0.0003 \\
        Update interval ($Q_\theta$) & 1 \\
        Update interval ($\pi_\phi$) & 2 \\
        Update interval (target $Q_\theta$) & 1 \\
        Initial entropy coefficient ($\alpha$) & 1.0 \\
        Automatic $\alpha$ tuning & True \\
        Activation & ReLU \\
        Optimizer & Adam \\
        \bottomrule
    \end{tabular}
\end{table}

\begin{table}[h!]
    \centering
    \caption{Task-specific discount factors for SAC and TD3.}
    \label{tab:gamma}
    \begin{tabular}{ll}
        \toprule
        \textbf{Task} & \textbf{Discount factor ($\gamma$)} \\
        \midrule
        \texttt{Hopper-v4} & 0.995 \\
        \texttt{Ant-v4} & 0.975 \\
        Others & 0.99 \\
        \bottomrule
    \end{tabular}
\end{table}

\paragraph{TD3.}
Our TD3 implementation also follows CleanRL~\citep{huang2022cleanrl}, with hyperparameters summarized in Table~\ref{tab:td3_hyperparameters} and discount factors specified in Table~\ref{tab:gamma}.
Actions are sampled uniformly during the initial 25K steps, followed by actor actions with Gaussian exploration noise.
Exploration and target policy noise are scaled by half the action-range width, with target noise clipped before scaling.
The resulting actions are clipped to the valid action bounds.
The actor and target networks are updated every two training steps, while both critics are updated every step.
We evaluate the deterministic policy without exploration noise every 10K steps, using one episode per checkpoint and reporting the original, uninflated return.

\begin{table}[h!]
    \centering
    \caption{TD3 hyperparameter configuration. Noise parameters are expressed relative to half the action-range width.}
    \label{tab:td3_hyperparameters}
    \begin{tabular}{ll}
        \toprule
        \textbf{Parameter} & \textbf{Value} \\
        \midrule
        Total environment steps & 3M \\
        Replay buffer size & 1M \\
        Batch size & 256 \\
        Warm-up steps & 25K \\
        Actor learning rate & 0.0003 \\
        Critic learning rate & 0.0003 \\
        Discount factor ($\gamma$) & See Table~\ref{tab:gamma} \\
        Target update coefficient ($\tau$) & 0.005 \\
        Critic update interval & 1 \\
        Actor update interval & 2 \\
        Target network update interval & 2 \\
        Exploration noise standard deviation & 0.2 \\
        Target policy noise standard deviation & 0.2 \\
        Target policy noise clipping & $[-0.5,\,0.5]$ \\
        Hidden layers (actor and critics) & $256,\,256$ \\
        Hidden activation & ReLU \\
        Optimizer & Adam \\
        Evaluation interval & 10K steps \\
        Evaluation episodes per checkpoint & 1 \\
        \bottomrule
    \end{tabular}
\end{table}

\paragraph{DQN.}
Our DQN implementation follows CleanRL~\citep{huang2022cleanrl}, with hyperparameters summarized in Table~\ref{tab:dqn_hyperparameters}.
ALE preprocessing includes up to 30 random no-op actions at reset, action repeat of four with max pooling, episodic-life handling, and fire-reset handling where applicable.
Sticky actions are disabled.
Observations are resized to $84\times84$ grayscale images and stacked over four frames.
Training rewards are sign-clipped before reward inflation is applied and stored in replay; reported episodic returns use the original, unclipped and uninflated environment rewards.
Step counts refer to interactions with the wrapped environment rather than individual emulator frames.

\begin{table}[h!]
    \centering
    \caption{DQN hyperparameter configuration for ALE. Convolutional specifications list channels, kernel size, and stride.}
    \label{tab:dqn_hyperparameters}
    \begin{tabular}{ll}
        \toprule
        \textbf{Parameter} & \textbf{Value} \\
        \midrule
        Total environment steps & 10M \\
        Replay buffer size & 1M \\
        Batch size & 32 \\
        Warm-up steps & 80K \\
        Learning rate & $10^{-4}$ \\
        Discount factor ($\gamma$) & 0.99 \\
        Gradient update interval & 4 \\
        Target network update interval & 1K \\
        Target update coefficient ($\tau$) & 1.0 (hard update) \\
        Initial exploration rate ($\epsilon$) & 1.0 \\
        Final exploration rate ($\epsilon$) & 0.01 \\
        Linear exploration decay duration & 1M steps \\
        Input shape & $4\times84\times84$ \\
        Input normalization & Divide by 255 \\
        Convolutional layer 1 & 32 channels, kernel 8, stride 4 \\
        Convolutional layer 2 & 64 channels, kernel 4, stride 2 \\
        Convolutional layer 3 & 64 channels, kernel 3, stride 1 \\
        Fully connected hidden layer & 512 units \\
        Output dimension & Number of actions \\
        Hidden activation & ReLU \\
        Optimizer & Adam \\
        \bottomrule
    \end{tabular}
\end{table}



\subsection{Fed Hyperparameters}
\label{app:fed_hyperparameters}

We selected the Fed hyperparameters through a simple search.
Table~\ref{tab:fed_hyperparameters} summarizes the configuration.

\begin{table}[h!]
    \centering
    \caption{Fed hyperparameter configuration.}
    \label{tab:fed_hyperparameters}
    \begin{tabular}{ll}
        \toprule
        \textbf{Parameter} & \textbf{Value} \\
        \midrule
        Short-term EMA coefficient ($\beta_s$) & 0 (instantaneous update) \\
        Long-term EMA coefficient ($\beta_l$) & 0.83 \\
        Inflation interval ($Y$) & $10^5$ environment steps \\
        Controller updates per interval ($K$) & 12 \\
        Target inflation rate ($\rho_Y^*$) & 0.02 \\
        Controller coefficient ($\lambda_1$) & 0.2 \\
        Controller coefficient ($\lambda_2$) & 0.2 \\
        Inflation-rate smoothing coefficient ($\tau_\rho$) & 0.83 \\
        \bottomrule
    \end{tabular}
\end{table}

\subsection{Implementation of Comparison Baselines}
\label{app:comparison_baselines}

\paragraph{Reward normalization (RN).}
We normalize rewards using the running standard deviation of observed rewards:
\[
    r_t^{\mathrm{RN}} = \frac{r_t}{\sigma_r}.
\]
We do not subtract the running mean. This avoids introducing an additive reward shift, which can alter incentives for episode termination.

\paragraph{Return normalization (RTN).}
Following the normalization form used by VecNormalize, we divide rewards by the running standard deviation of discounted returns:
\[
    r_t^{\mathrm{RTN}} = \frac{r_t}{\sigma_R}.
\]
Both RN and RTN use running statistics rather than fixed statistics. Automatic entropy-temperature tuning remains enabled in all SAC comparisons.

\paragraph{Learning-rate inflation.}
The ``2\% (lr)'' baseline applies the same multiplicative schedule as 2\% reward inflation to the learning rates instead of rewards. Thus, the learning rates increase throughout training, rather than remaining at a fixed larger value.

\paragraph{Prioritized experience replay (PER).}
We use proportional prioritized experience replay~\citep{schaul2015prioritized}, sampling transition $i$ with probability $P(i)=p_i^{0.6}/\sum_j p_j^{0.6}$. Priorities are updated as $p_i=|\delta_i|+10^{-6}$, where $\delta_i$ is the first critic's TD error computed using the SAC target. New transitions receive the maximum priority currently in the buffer. Both critic losses use importance-sampling weights $w_i=(N P(i))^{-\beta}$, normalized by the maximum weight within the sampled minibatch, where $N$ is the current buffer size. We initialize $\beta=0.4$ and increase it by $10^{-6}$ after each sampling call, capped at $1$. Actor and entropy-temperature losses use the same sampled transitions without importance weighting.

\paragraph{Recency-weighted sampling (RWS).}
We designed RWS using a recency-biased replay buffer, where samples are drawn according to the probability mass function:
$$p(a)=\frac{(1-r)r^{a-1}}{1-r^N},$$
where $a=1,2,\dots,N$ denotes the sample age, $N$ is the replay buffer size, and $0<r<1$ is the decay rate.
This assigns higher probability to more recent transitions (smaller $a$), with the ratio $\frac{p(a_2)}{p(a_1)} = r^{a_2-a_1}$ for $a_2>a_1$.
Thus, the sampling scheme imposes a geometric decay over sample age, analogous to the geometric scaling of rewards across global steps induced by reward inflation.
We set $r = 0.98^{1/100000}$ to align with the notion of $2\%$ reward inflation.

RWS biases the sampling process, causing the networks ($Q_\theta$ or $\pi_\phi$) to encounter recent transitions more frequently.
In contrast, reward inflation preserves uniform sampling but assigns relatively larger rewards to recently acquired transitions.

\paragraph{Plasticity baselines.}
For ReDo, we use a recycling period of 1K steps and a dormant-neuron threshold of 0.025. Reset is applied every 200K steps. Shrink \& Perturb is applied every 1K steps with a shrink factor of 0.99.

\subsection{Tabular Q-Learning Settings}
\label{app:tabular_settings}

\begin{figure}[t]
    \centering
    \includegraphics[width=0.8\linewidth]{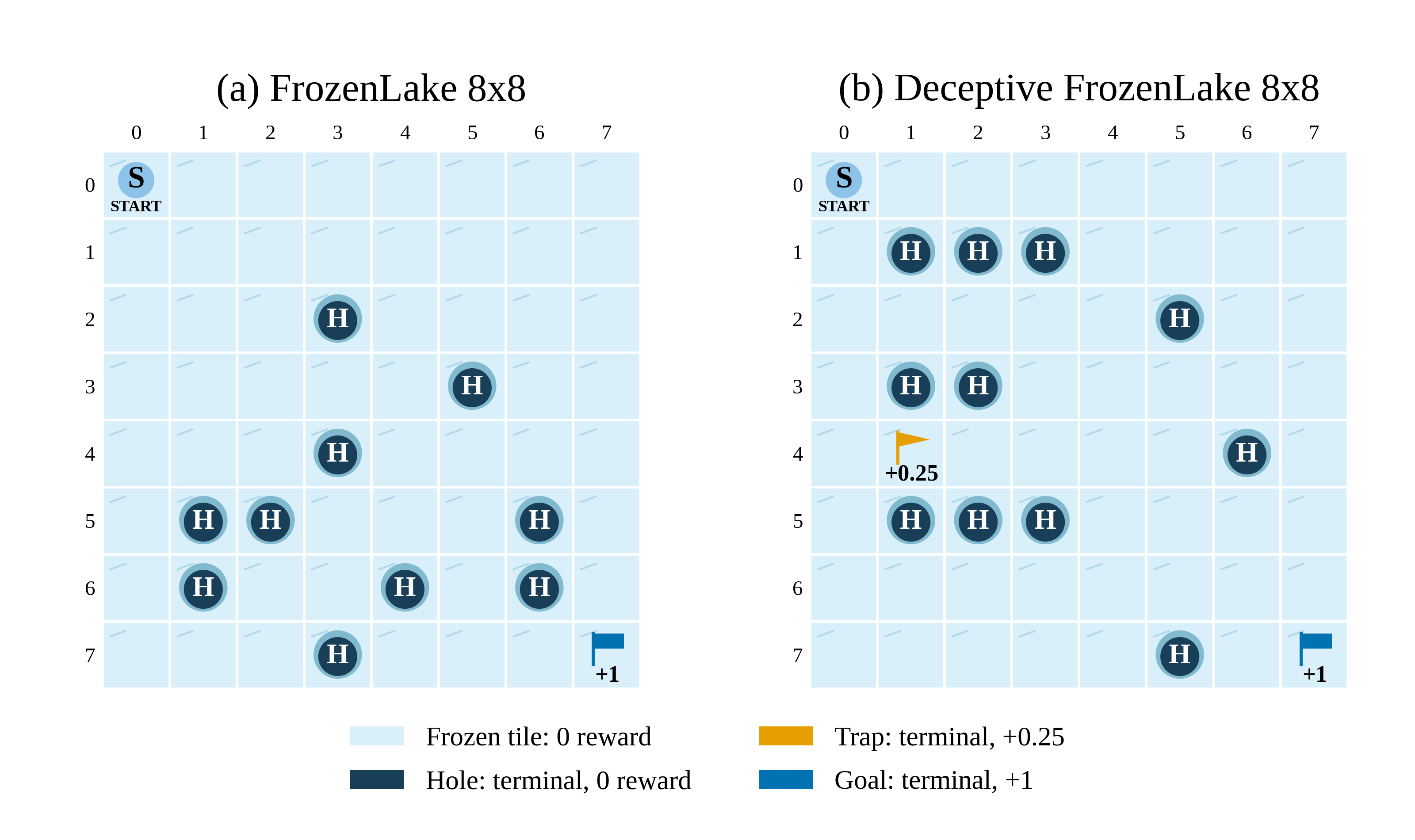}
    \caption{FrozenLake maps: (a) the standard map and (b) the deceptive-reward map, with terminal goal rewards of 0.25 and 1.0. Both environments use slippery transitions.}
    \label{fig:frozenlake_maps}
\end{figure}

\paragraph{Environments.}
Figure~\ref{fig:frozenlake_maps} shows the two $8\times8$ FrozenLake maps.
The standard setting uses the slippery \texttt{FrozenLake-v1} environment.
The deceptive-reward setting also uses slippery transitions, with a start at $(0,0)$, a low-reward goal at $(4,1)$ yielding 0.25, and a high-reward goal at $(7,7)$ yielding 1.0, using zero-based row and column indices.
All other transitions yield zero reward.
Reaching either goal or falling into a hole terminates the episode; episodes are truncated after 200 steps.
The deceptive-reward experiment shares the replay Q-learning settings in Table~\ref{tab:tabular_hyperparameters}, with $\epsilon$ reaching 0.01 at 1M steps under the lower-exploration schedule and at 2M steps under the higher-exploration schedule.
We classify a run as trapped if its final greedy policy reaches the low-reward goal more frequently than the high-reward goal over 1,000 evaluation episodes under slippery dynamics.
The reported percentage is the fraction of the 32 runs classified as trapped.

\paragraph{Algorithm and hyperparameters.}
We use tabular Q-learning with a zero-initialized Q-table and uniform experience replay.
After collecting 100 transitions, we perform one update per environment step using a single transition sampled uniformly from the replay buffer.
Rewards are multiplied by the collection-time inflation level before insertion and remain unchanged thereafter.
Bootstrapping is disabled for transitions ending in either termination or truncation.
Table~\ref{tab:tabular_hyperparameters} summarizes the settings.
The $\epsilon$-greedy exploration rate decreases linearly from 1.0 to 0.01 over the first 1M steps and remains at 0.01 thereafter.
Only the higher-exploration condition in the deceptive-reward experiment extends this decay to the full 2M training steps.
For the standard-map experiment, greedy policies are evaluated every 10K steps over 200 episodes, with a separate final evaluation over 1,000 episodes.
Success rate is the fraction of evaluation episodes reaching the goal.
Learning-curve AUC is computed by trapezoidal integration and divided by the training horizon.
For the recency analysis, we record $\log(|\delta|+10^{-12})$ before each replay update, aggregate it by transition age for reward-bearing transitions, and fit weighted least squares using the number of samples at each age as the weight.

\begin{table}[h!]
    \centering
    \caption{Tabular Q-learning hyperparameters. The longer exploration decay applies only to the higher-exploration condition in the deceptive-reward experiment.}
    \label{tab:tabular_hyperparameters}
    \begin{tabular}{ll}
        \toprule
        \textbf{Parameter} & \textbf{Value} \\
        \midrule
        Total environment steps & 2M \\
        Learning rate & 0.1 \\
        Discount factor ($\gamma$) & 0.99 \\
        Initial Q-values & 0 \\
        Replay buffer capacity & 10K \\
        Warm-up transitions & 100 \\
        Replay updates per environment step & 1 \\
        Transitions per replay update & 1 \\
        Initial exploration rate ($\epsilon$) & 1.0 \\
        Final exploration rate ($\epsilon$) & 0.01 \\
        Linear exploration decay duration & 1M steps \\
        Higher-exploration decay duration & 2M steps \\
        Inflation interval ($Y$) & $10^5$ steps \\
        \bottomrule
    \end{tabular}
\end{table}

\subsection{Resources}
\label{subsect:resources}
All experiments were conducted on a system with multiple Linux servers including GPUs as NVIDIA RTX 4090, RTX A6000, RTX A5000, and TITAN RTX, paired with CPUs as AMD EPYC 9354 32-Core, AMD EPYC 9554 64-Core, AMD Ryzen Threadripper PRO 5975WX 32-Cores and AMD Ryzen Threadripper 3970X 32-Cores, respectively. On average, a single experimental run for both ALE (10M steps) and MuJoCo (3M steps) takes approximately 8 hours on the reported hardware.

\section{Additional Results}

\begin{table}[h!]
\centering
\caption{Performance comparison of DQN agents in ALE games under reward inflation (0\%, 2\%, and Fed (2\%)).}
\small
\setlength{\tabcolsep}{0pt}
\begin{tabular*}{\linewidth}{@{\extracolsep{\fill}} l r r r l r r r @{}}
\toprule
\textbf{Task} & \textbf{0\%} & \textbf{2\%} & \textbf{Fed (2\%)} &
\textbf{Task} & \textbf{0\%} & \textbf{2\%} & \textbf{Fed (2\%)} \\
\midrule

Alien & 1250{\scriptsize $\pm$352} & \textbf{1513}{\scriptsize $\pm$124} & 1482{\scriptsize $\pm$96} &
Frostbite & 2732{\scriptsize $\pm$364} & \textbf{3377}{\scriptsize $\pm$543} & 3172{\scriptsize $\pm$579} \\

Amidar & 296{\scriptsize $\pm$33} & \textbf{468}{\scriptsize $\pm$64} & 456{\scriptsize $\pm$83} &
Gopher & 12595{\scriptsize $\pm$3870} & \textbf{16160}{\scriptsize $\pm$4948} & 15954{\scriptsize $\pm$5352} \\

Assault & 3654{\scriptsize $\pm$252} & \textbf{3814}{\scriptsize $\pm$449} & 3401{\scriptsize $\pm$396} &
Hero & \textbf{2758}{\scriptsize $\pm$1779} & 2731{\scriptsize $\pm$624} & 2549{\scriptsize $\pm$215} \\

Asterix & 22178{\scriptsize $\pm$7214} & \textbf{31424}{\scriptsize $\pm$5245} & 29492{\scriptsize $\pm$3913} &
Jamesbond & 486{\scriptsize $\pm$63} & \textbf{601}{\scriptsize $\pm$154} & 567{\scriptsize $\pm$120} \\

BankHeist & 650{\scriptsize $\pm$251} & 776{\scriptsize $\pm$189} & \textbf{797}{\scriptsize $\pm$109} &
Kangaroo & 7821{\scriptsize $\pm$1762} & 10885{\scriptsize $\pm$2742} & \textbf{11983}{\scriptsize $\pm$2016} \\

BattleZone & 22659{\scriptsize $\pm$1074} & 22195{\scriptsize $\pm$4406} & \textbf{24501}{\scriptsize $\pm$945} &
Krull & 7584{\scriptsize $\pm$2605} & 7969{\scriptsize $\pm$2602} & \textbf{8799}{\scriptsize $\pm$419} \\

BeamRider & 6906{\scriptsize $\pm$662} & \textbf{8682}{\scriptsize $\pm$904} & 7908{\scriptsize $\pm$759} &
KungFu & 3853{\scriptsize $\pm$4542} & 8257{\scriptsize $\pm$8977} & \textbf{12075}{\scriptsize $\pm$11337} \\

Bowling & 22.46{\scriptsize $\pm$0.63} & \textbf{23.55}{\scriptsize $\pm$0.90} & 23.30{\scriptsize $\pm$0.50} &
MsPacman & 2218{\scriptsize $\pm$141} & \textbf{2365}{\scriptsize $\pm$102} & 2181{\scriptsize $\pm$135} \\

Boxing & 91.37{\scriptsize $\pm$2.06} & 94.93{\scriptsize $\pm$1.02} & \textbf{95.44}{\scriptsize $\pm$1.00} &
Pong & 20.40{\scriptsize $\pm$0.1} & \textbf{20.58}{\scriptsize $\pm$0.09} & 20.30{\scriptsize $\pm$0.40} \\

Breakout & 367{\scriptsize $\pm$19} & \textbf{375}{\scriptsize $\pm$14} & \textbf{375}{\scriptsize $\pm$21} &
PrivateEye & \textbf{621}{\scriptsize $\pm$1229} & 63{\scriptsize $\pm$117} & 81{\scriptsize $\pm$143} \\

ChopperCmd & 1178{\scriptsize $\pm$508} & 1492{\scriptsize $\pm$658} & \textbf{1553}{\scriptsize $\pm$1301} &
Qbert & 11754{\scriptsize $\pm$1849} & 12317{\scriptsize $\pm$1376} & \textbf{12831}{\scriptsize $\pm$1756} \\

CrazyClimber & 93919{\scriptsize $\pm$8113} & 103214{\scriptsize $\pm$10735} & \textbf{105703}{\scriptsize $\pm$10671} &
Riverraid & 8228{\scriptsize $\pm$641} & \textbf{8699}{\scriptsize $\pm$1122} & 8350{\scriptsize $\pm$840} \\

DemonAttack & 32572{\scriptsize $\pm$9621} & 50007{\scriptsize $\pm$6358} & \textbf{50466}{\scriptsize $\pm$6846} &
RoadRunner & 34017{\scriptsize $\pm$1172} & 35502{\scriptsize $\pm$6092} & \textbf{40112}{\scriptsize $\pm$3054} \\

Enduro & 1275{\scriptsize $\pm$108} & 1542{\scriptsize $\pm$165} & \textbf{1553}{\scriptsize $\pm$150} &
Seaquest & 4074{\scriptsize $\pm$632} & \textbf{5138}{\scriptsize $\pm$1024} & 4635{\scriptsize $\pm$1048} \\

Freeway & 23.44{\scriptsize $\pm$3.58} & \textbf{24.81}{\scriptsize $\pm$4.80} & 23.53{\scriptsize $\pm$3.73} &
SpaceInvade & 1349{\scriptsize $\pm$106} & \textbf{1565}{\scriptsize $\pm$148} & 1483{\scriptsize $\pm$222} \\

\bottomrule
\end{tabular*}
\label{tab:atari_inflation_full}
\end{table}

\begin{figure*}
  \centering
  \includegraphics[width=0.9\linewidth]{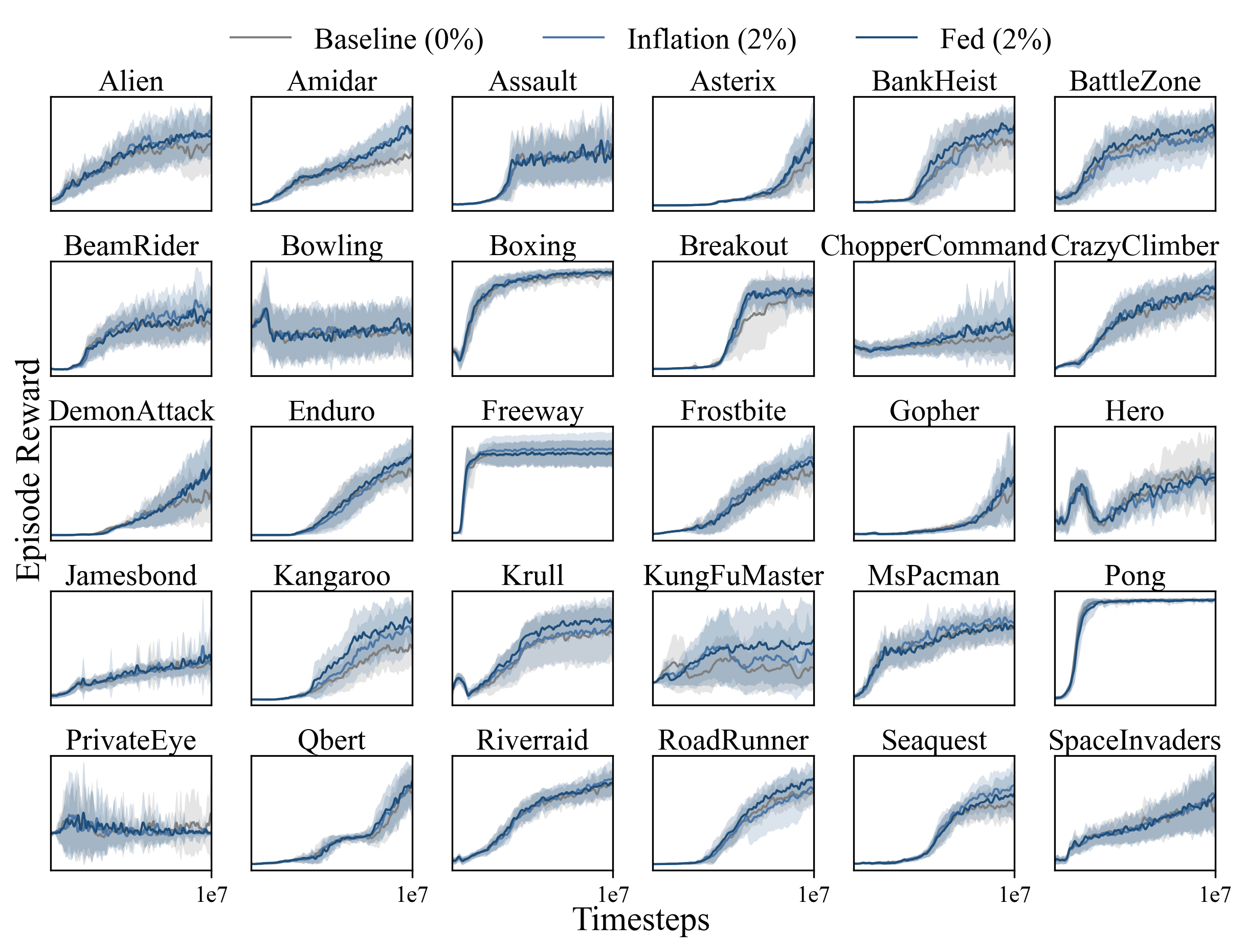}
  \caption{Learning curves of DQN agents across all ALE games under different reward inflation settings ($0\%$, $2\%$, and Fed with $2\%$ target).}
  \label{fig:atari_learning_curves}
\end{figure*}

\begin{figure}[t]
    \centering
    \includegraphics[width=0.4\textwidth]{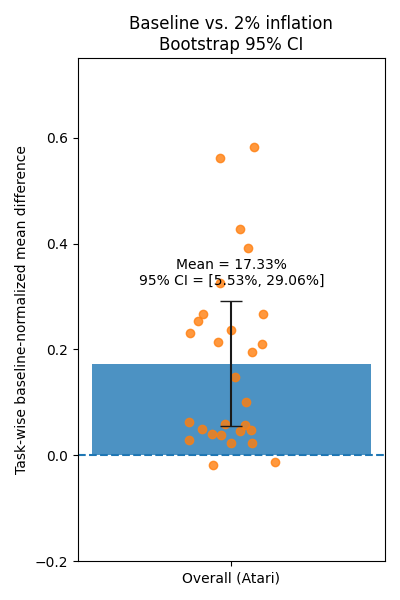}
    \includegraphics[width=0.4\textwidth]{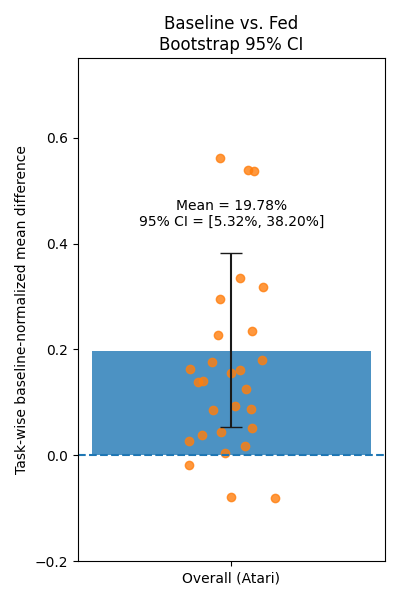}
    \label{fig:combined}
    \caption{Task-wise bootstrap 95\% CI over Atari tasks. (Left) Baseline vs. 2\% inflation rate. (Right) Baseline vs. Fed.}
\label{fig:bootstrapCI_taskwise_atari}
\end{figure}

\begin{figure}[t]
    \centering
    \includegraphics[width=0.4\textwidth]{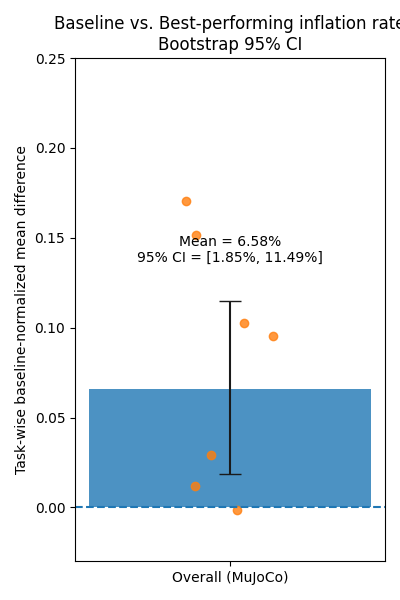}
    \includegraphics[width=0.4\textwidth]{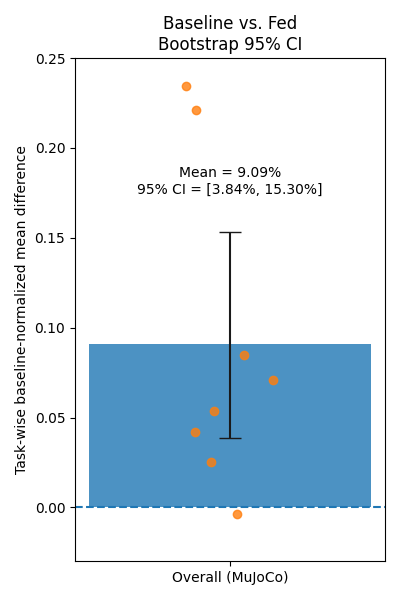}
    \label{fig:combined}
    \caption{Task-wise bootstrap 95\% CI over MuJoCo tasks. (Left) Baseline vs. the best-performing inflation rate identified in Table 1 of the main paper. (Right) Baseline vs. Fed.}
\label{fig:bootstrapCI_taskwise_mujoco}
\end{figure}

\begin{table}[t]
    \caption{TD3 performance on eight MuJoCo tasks with and without 2\% reward inflation. Results are mean $\pm$ standard deviation across eight runs trained for 3M timesteps. Final scores are averaged over the final 10\% of training, and AUC is computed over the full learning curve. All evaluations use the original, uninflated rewards. Bold indicates the higher mean within each metric and task.}
    \label{tab:td3_results}
    \centering
    \small
    \setlength{\tabcolsep}{4pt}
    \begin{tabular}{lcccc}
        \toprule
        & \multicolumn{2}{c}{\textbf{Final score}}
        & \multicolumn{2}{c}{\textbf{AUC}} \\
        \cmidrule(lr){2-3}
        \cmidrule(lr){4-5}
        \textbf{Task} & \textbf{0\%} & \textbf{2\%}
        & \textbf{0\%} & \textbf{2\%} \\
        \midrule
        HalfCheetah
        & $13662 \pm 651$ & $\mathbf{13730} \pm 618$
        & $11213 \pm 592$ & $\mathbf{11537} \pm 220$ \\
        Walker2d
        & $4862 \pm 471$ & $\mathbf{4881} \pm 557$
        & $3858 \pm 443$ & $\mathbf{3889} \pm 773$ \\
        Humanoid
        & $5788 \pm 338$ & $\mathbf{6023} \pm 503$
        & $4854 \pm 222$ & $\mathbf{5010} \pm 279$ \\
        Swimmer
        & $\mathbf{122} \pm 25$ & $120 \pm 37$
        & $94 \pm 23$ & $\mathbf{97} \pm 27$ \\
        Hopper
        & $3075 \pm 397$ & $\mathbf{3154} \pm 383$
        & $2650 \pm 330$ & $\mathbf{2796} \pm 148$ \\
        Ant
        & $\mathbf{6540} \pm 224$ & $6207 \pm 255$
        & $5192 \pm 284$ & $\mathbf{5212} \pm 202$ \\
        Pusher
        & $-23.69 \pm 3.66$ & $\mathbf{-22.45} \pm 1.73$
        & $-25.45 \pm 1.54$ & $\mathbf{-25.33} \pm 1.28$ \\
        Reacher
        & $-3.67 \pm 0.21$ & $\mathbf{-3.66} \pm 0.23$
        & $\mathbf{-3.73} \pm 0.08$ & $-3.74 \pm 0.10$ \\
        \bottomrule
    \end{tabular}
\end{table}

\begin{figure}[t]
    \centering
    \includegraphics[width=0.4\textwidth]{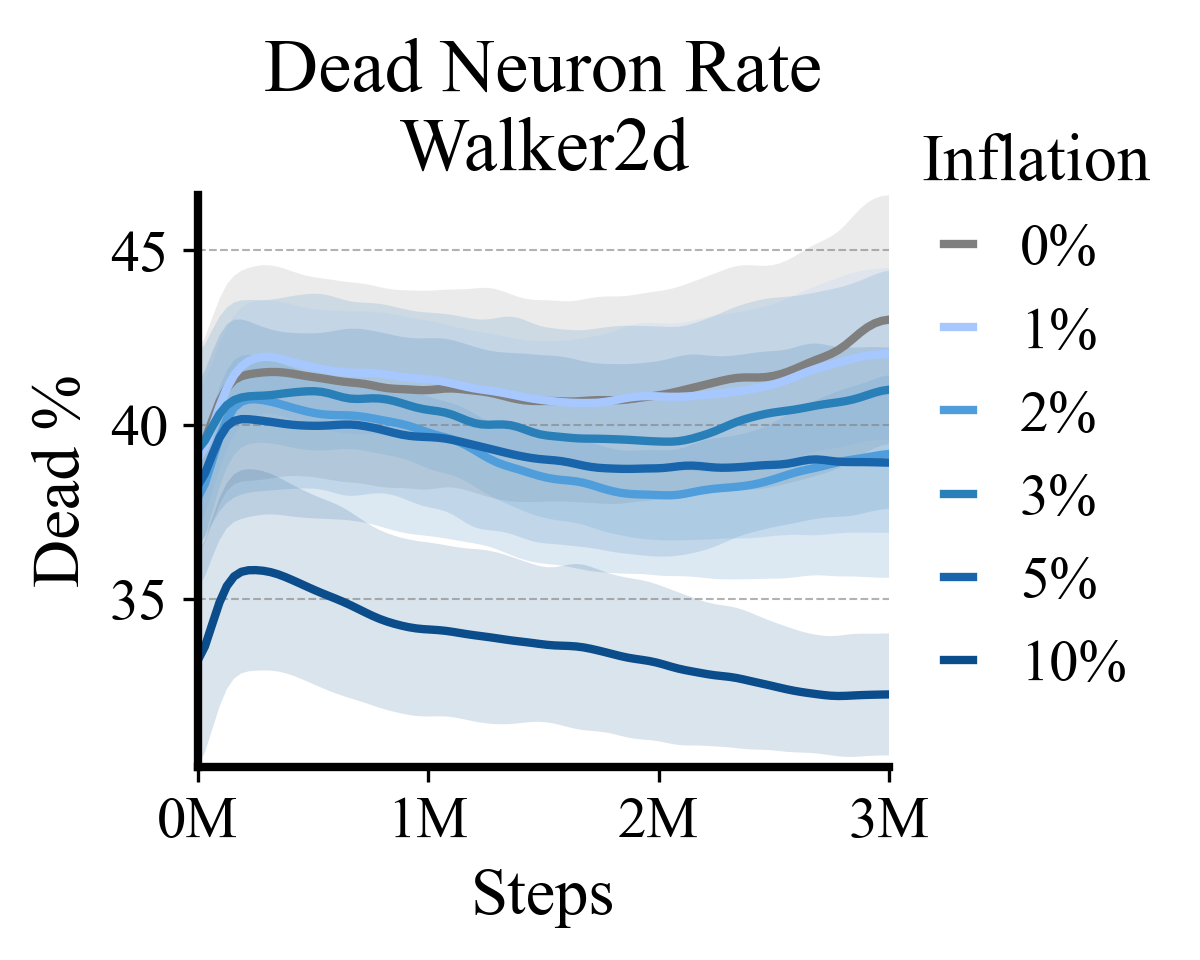}
    \includegraphics[width=0.4\textwidth]{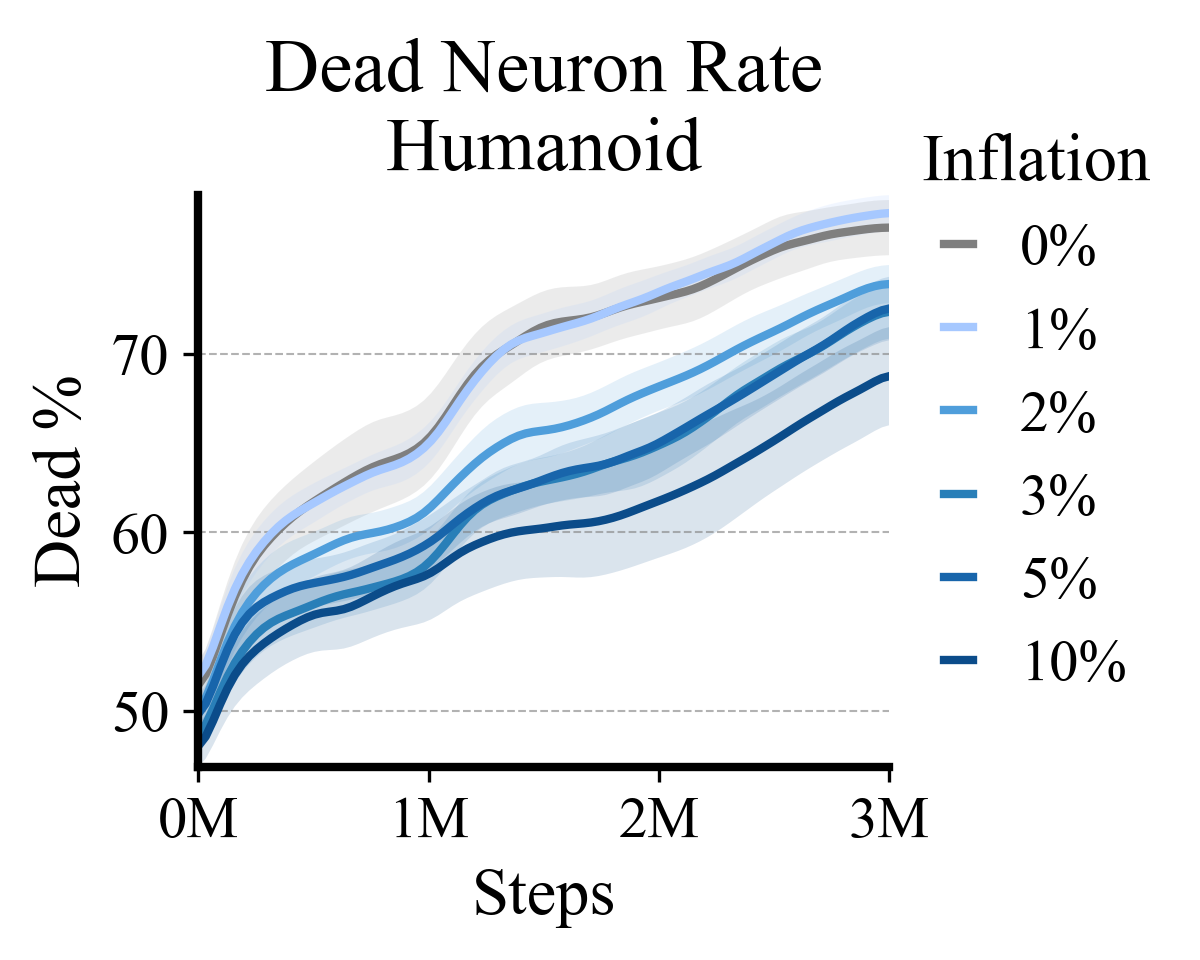}
    \includegraphics[width=0.4\textwidth]{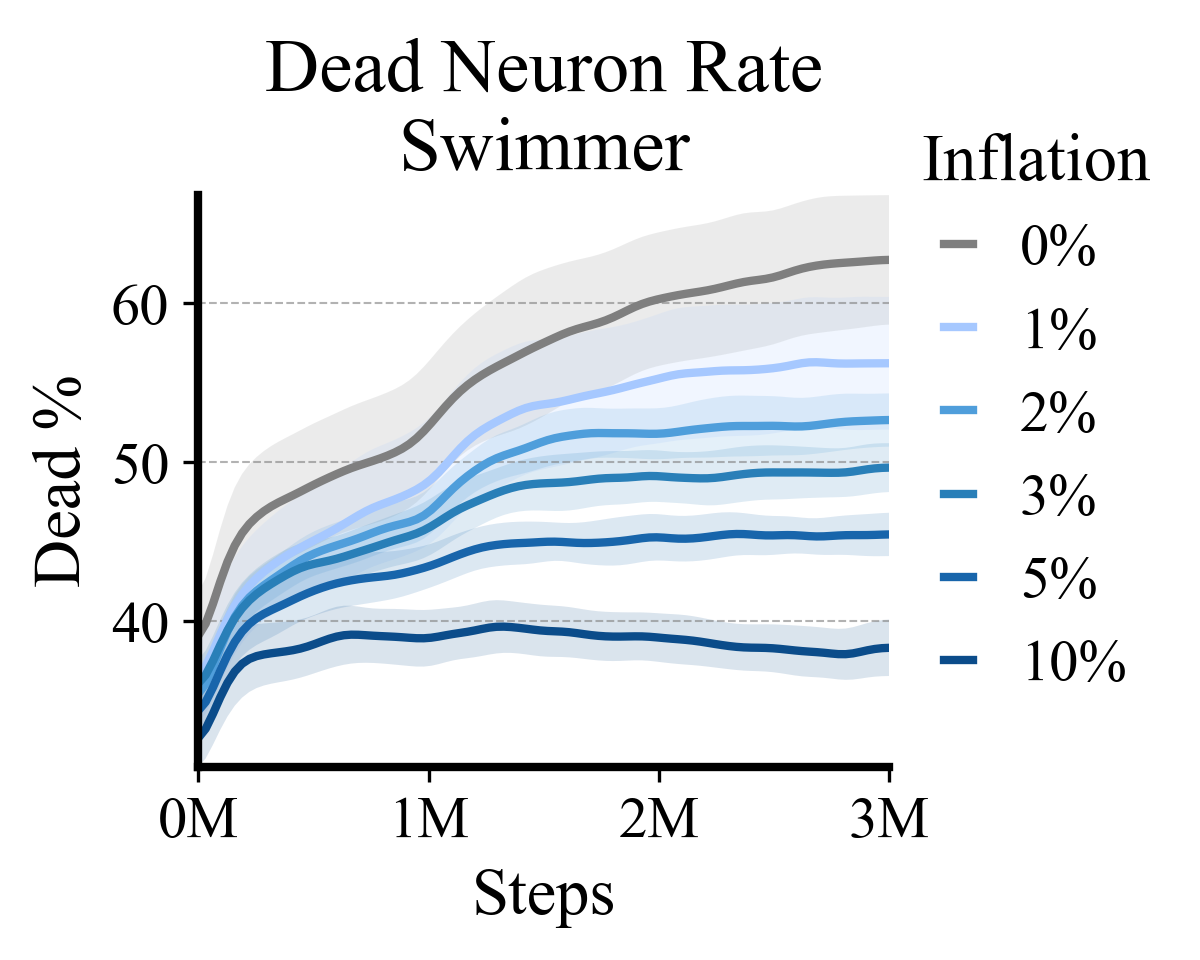}
    \label{fig:combined}
    \caption{Additional results of dormant (dead) neuron ratios across different tasks under varying reward inflation rates (Fig. 2(a) in the main paper corresponds to HalfCheetah-v4).}
\label{fig:dead_relu_additional}
\end{figure}

\begin{figure}[t]
    \centering
    \includegraphics[width=0.4\textwidth]{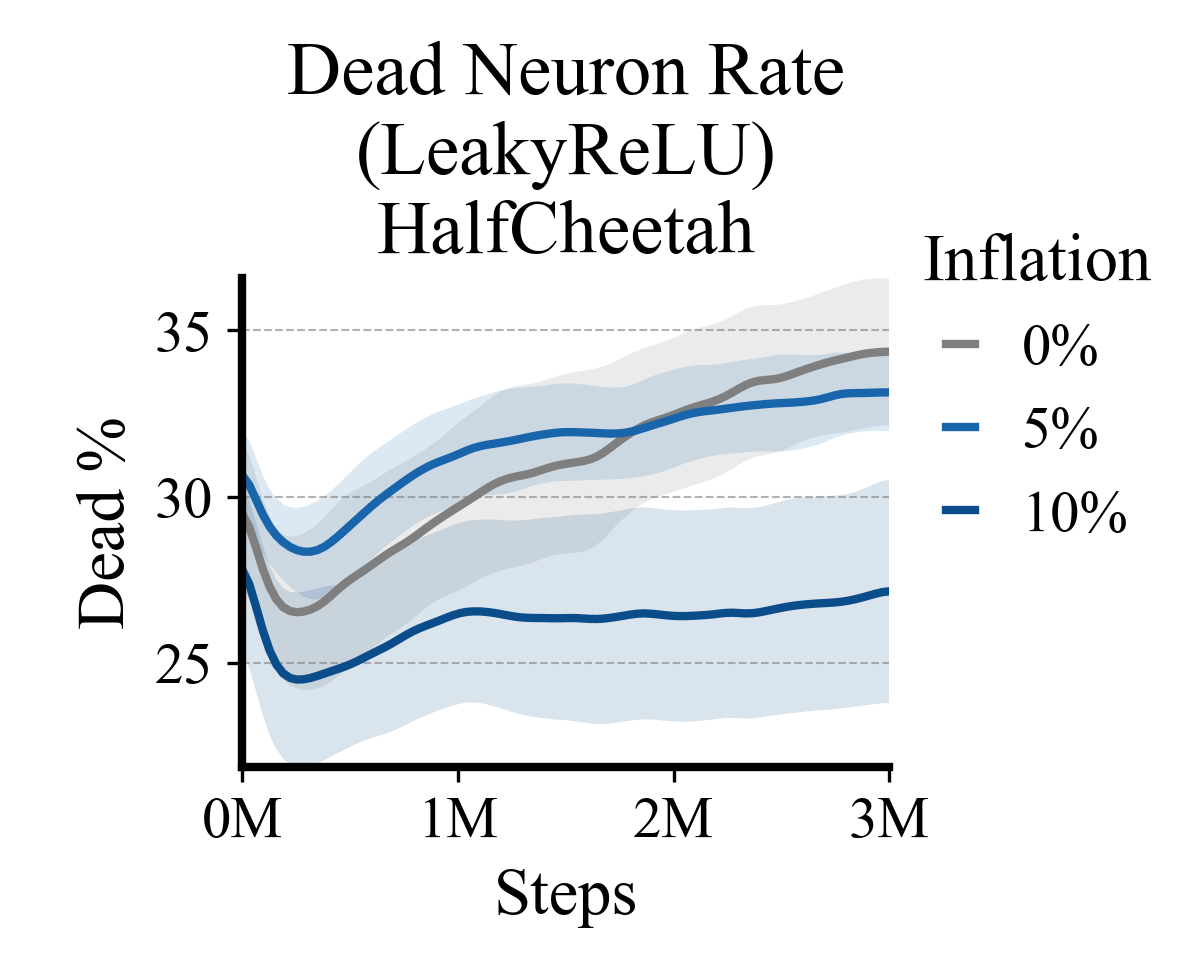}
    \includegraphics[width=0.4\textwidth]{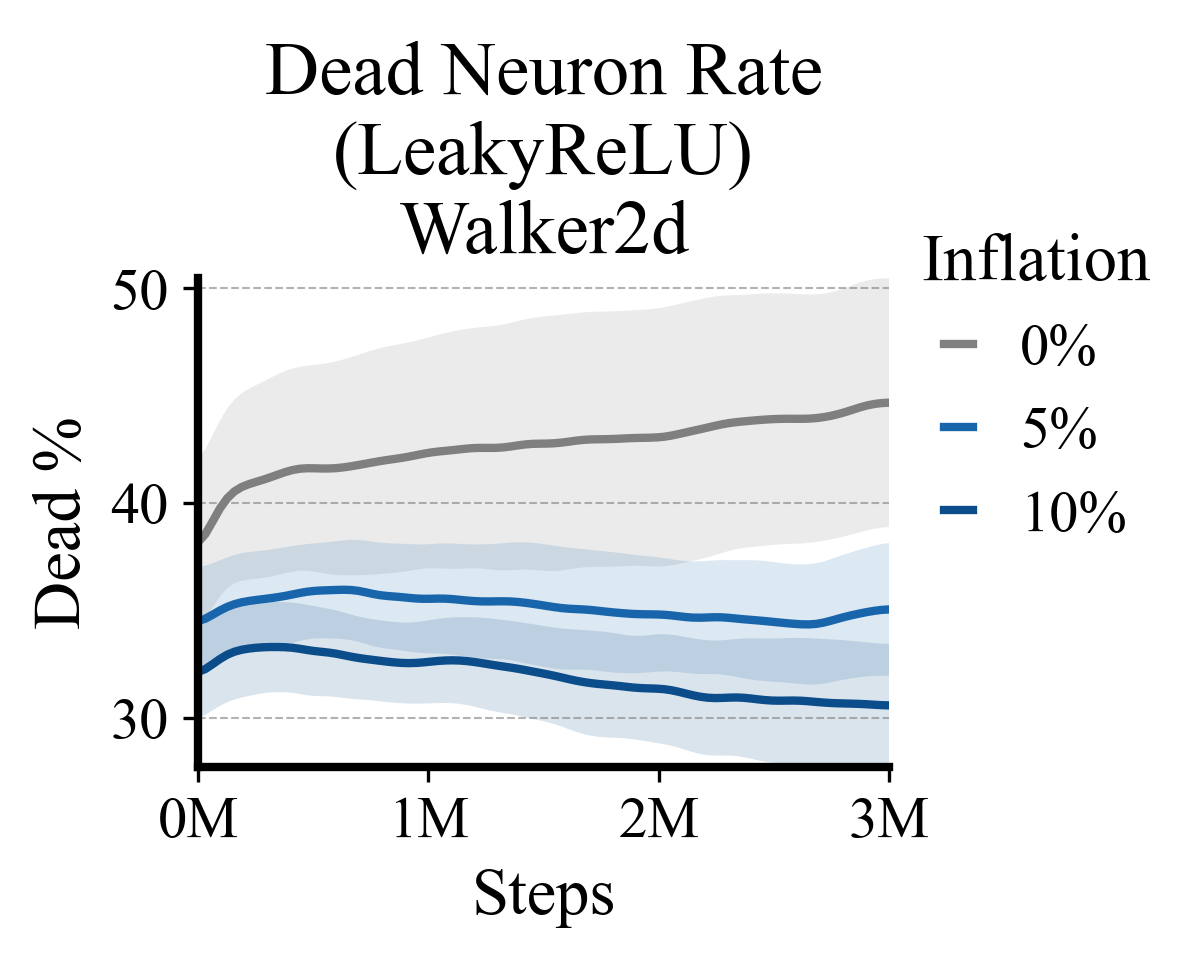}
    \includegraphics[width=0.4\textwidth]{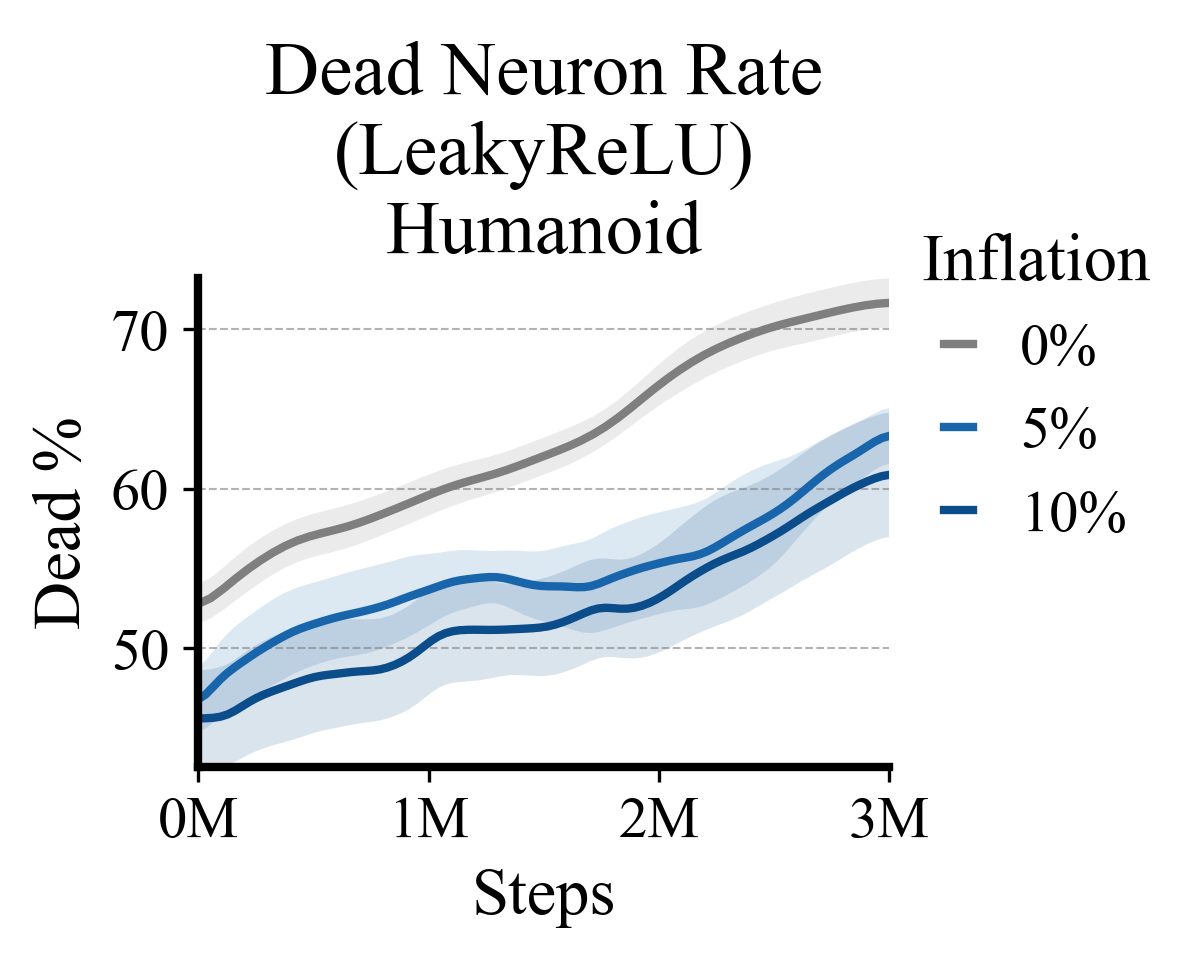}
    \includegraphics[width=0.4\textwidth]{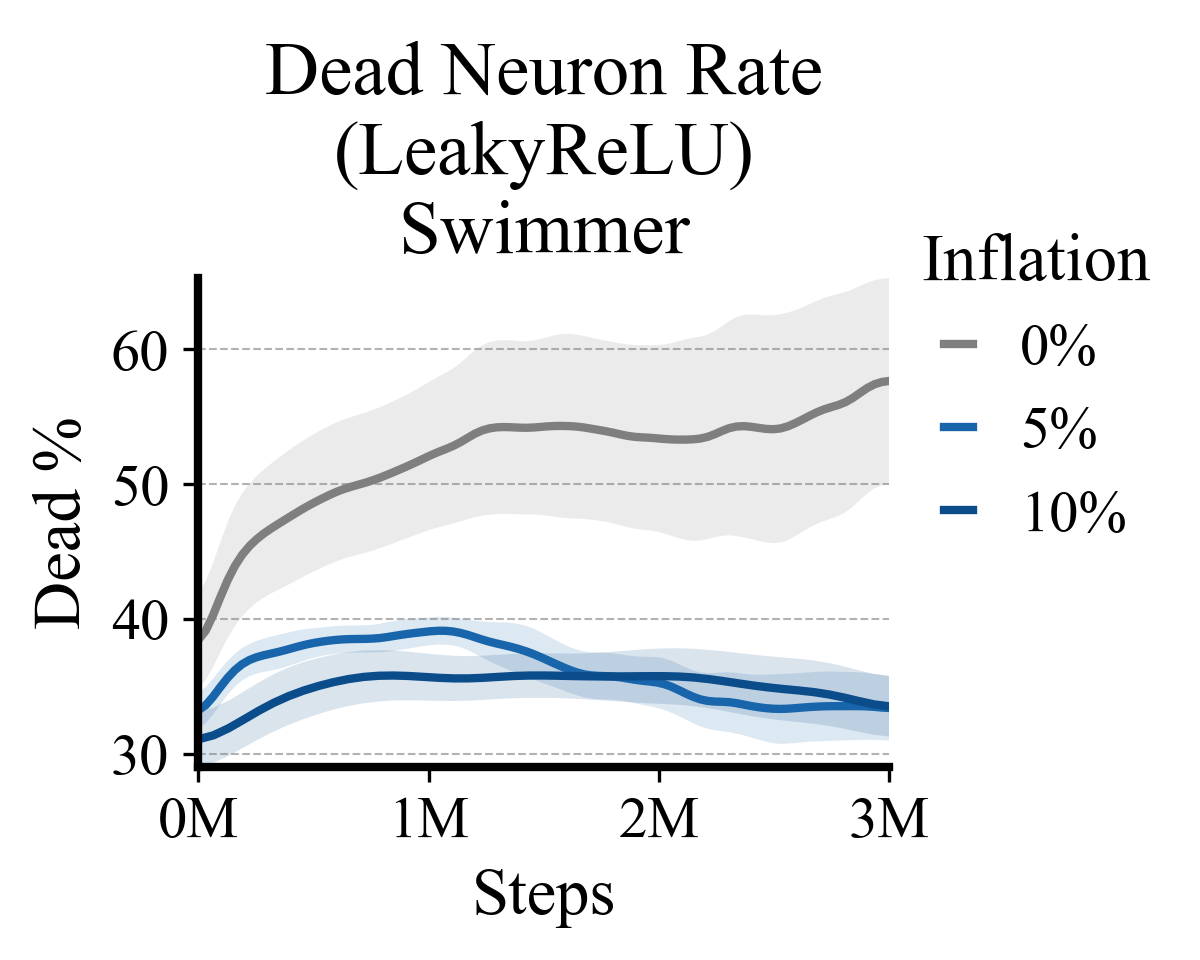}
    \label{fig:combined}
    \caption{Dormant (dead) neuron ratios across different tasks under varying reward inflation rates, using \texttt{LeakyReLU} instead of \texttt{ReLU}. Neurons whose negative activation ratio exceeded 99\% across inputs were counted as dormant.}
\label{fig:dead_leakyrelu}
\end{figure}

\begin{figure}[t]
    \centering
    \includegraphics[width=0.4\textwidth]{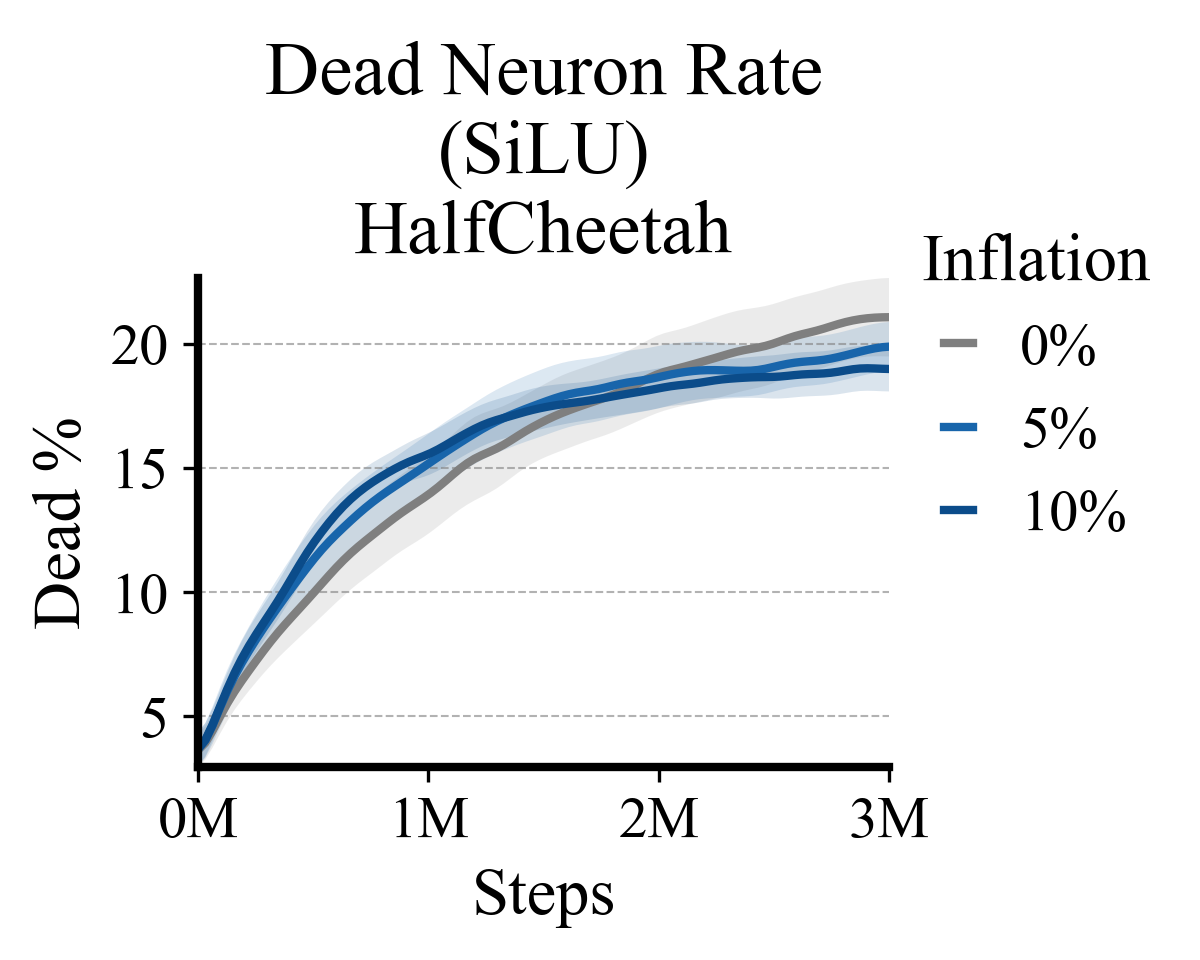}
    \includegraphics[width=0.4\textwidth]{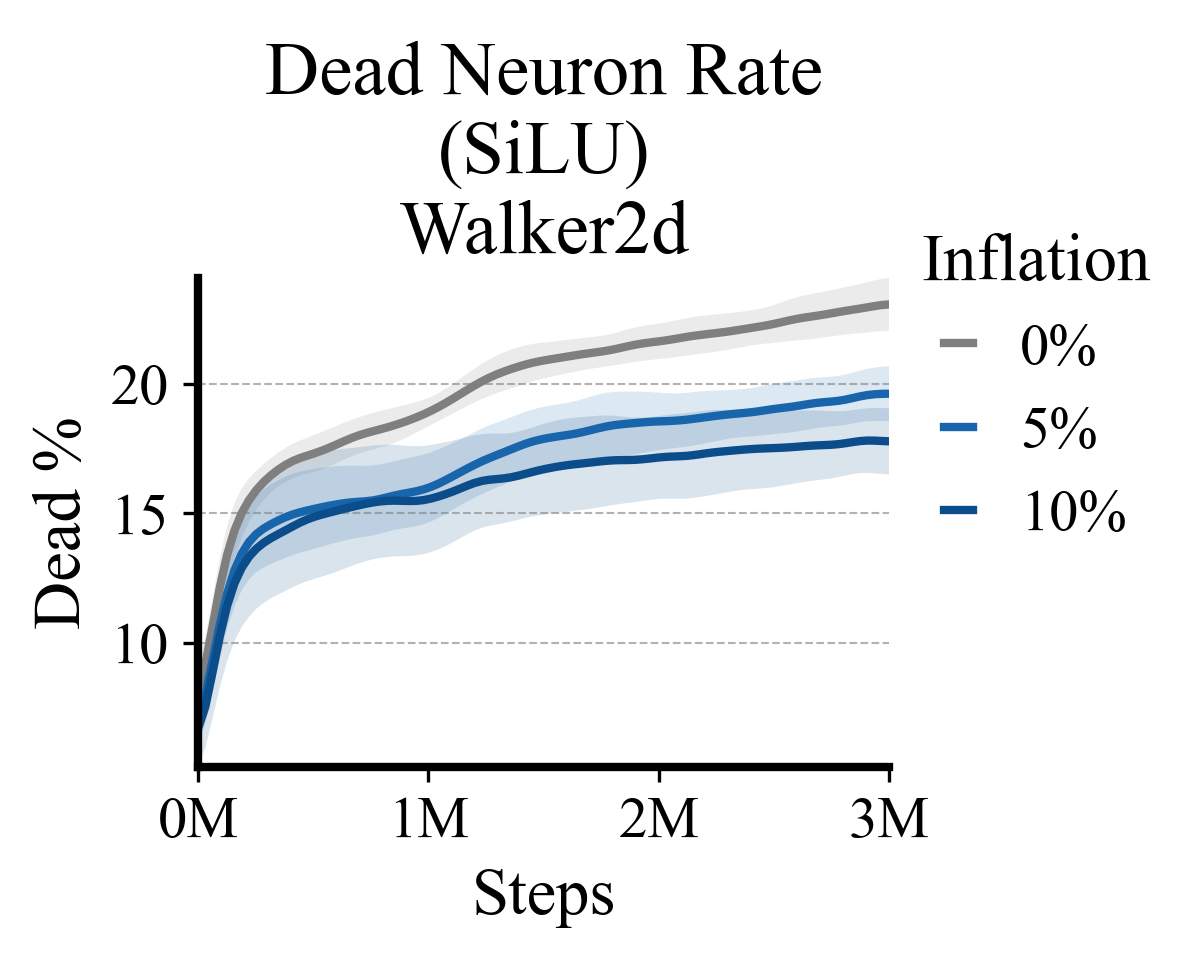}
    \includegraphics[width=0.4\textwidth]{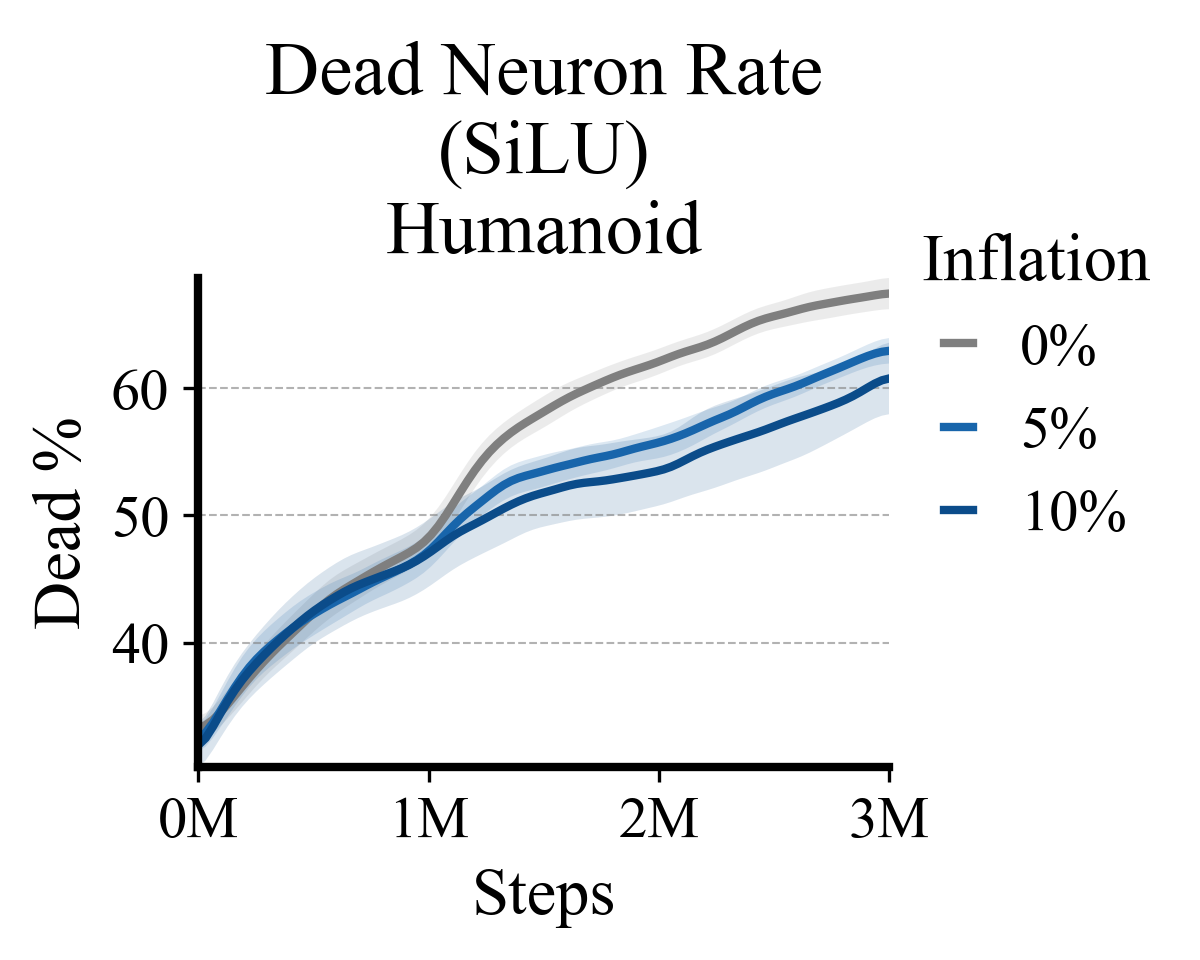}
    \includegraphics[width=0.4\textwidth]{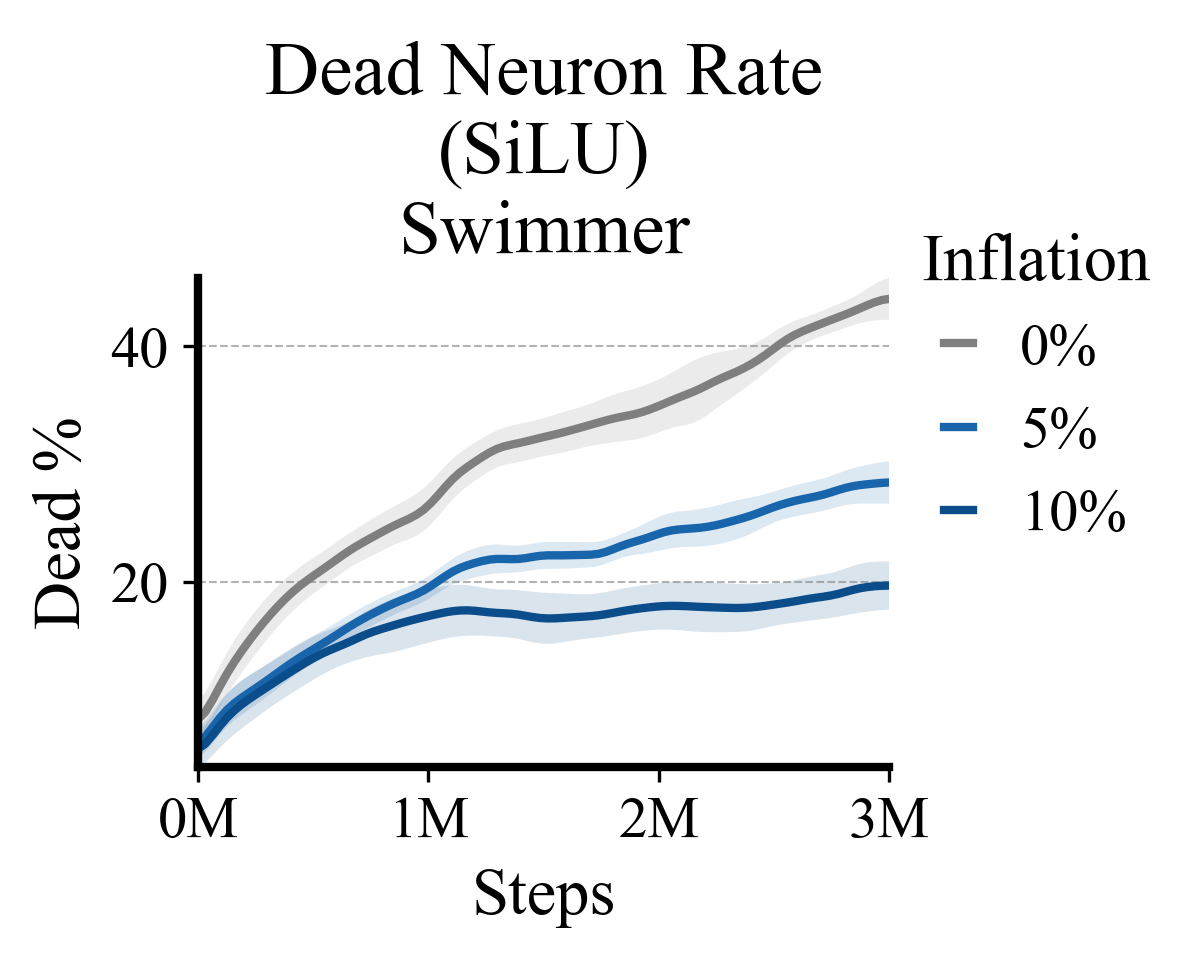}
    \label{fig:combined}
    \caption{Dormant (dead) neuron ratios across different tasks under varying reward inflation rates, using \texttt{SiLU} instead of \texttt{ReLU}. Neurons whose activation variance across inputs was below 0.0001 were counted as dormant.}
\label{fig:dead_silu}
\end{figure}

\begin{table}[t]
    \caption{Performance comparison with plasticity baselines on MuJoCo. Results are mean $\pm$ standard deviation of returns under the original, uninflated rewards, averaged over the final 300K training steps unless otherwise indicated. Reset (100K) and Reset (10K) use shorter evaluation windows, not different reset intervals. Bold indicates the highest mean in each task.}
    \label{tab:plasticity_baselines}
    \centering
    \small
    \setlength{\tabcolsep}{0pt}
    \resizebox{\linewidth}{!}{
    \begin{tabular}{lcccccccc}
        \toprule
        \textbf{Method}
        & \textbf{HalfCheetah} & \textbf{Walker2d}
        & \textbf{Humanoid} & \textbf{Swimmer}
        & \textbf{Hopper} & \textbf{Ant}
        & \textbf{Pusher} & \textbf{Reacher} \\
        \midrule
        0\%
        & $15241 \pm 734$ & $4399 \pm 527$
        & $5826 \pm 313$ & $116 \pm 36$
        & $2869 \pm 299$ & $4763 \pm 1642$
        & $-23.28 \pm 2.89$ & $\mathbf{-3.41} \pm 0.23$ \\
        2\%
        & $15607 \pm 372$ & $5102 \pm 520$
        & $5861 \pm 199$ & $136 \pm 11$
        & $3040 \pm 499$ & $4346 \pm 1267$
        & $-22.07 \pm 1.58$ & $-3.42 \pm 0.24$ \\
        Fed (2\%)
        & $\mathbf{15621} \pm 991$ & $\mathbf{5395} \pm 452$
        & $\mathbf{6048} \pm 436$ & $\mathbf{143} \pm 13$
        & $3105 \pm 293$ & $\mathbf{5180} \pm 1352$
        & $-21.99 \pm 1.23$ & $-3.43 \pm 0.24$ \\
        \midrule
        ReDo
        & $13692 \pm 1933$ & $5223 \pm 652$
        & $5494 \pm 247$ & $108 \pm 35$
        & $\mathbf{3380} \pm 122$ & $4986 \pm 1231$
        & $\mathbf{-21.58} \pm 0.63$ & $-3.46 \pm 0.24$ \\
        Reset
        & $8823 \pm 257$ & $3379 \pm 373$
        & $3486 \pm 301$ & $71 \pm 33$
        & $2555 \pm 170$ & $1919 \pm 1093$
        & $-26.61 \pm 1.81$ & $-4.19 \pm 0.54$ \\
        \quad (100K)
        & $10915 \pm 649$ & $3862 \pm 505$
        & $4531 \pm 462$ & $77 \pm 37$
        & $2892 \pm 238$ & $2397 \pm 1553$
        & $-24.38 \pm 3.76$ & $-3.87 \pm 0.58$ \\
        \quad (10K)
        & $11661 \pm 517$ & $3936 \pm 1813$
        & $5351 \pm 694$ & $78 \pm 38$
        & $3199 \pm 748$ & $2746 \pm 1618$
        & $-23.11 \pm 3.79$ & $-3.85 \pm 1.13$ \\
        S\&P
        & $10936 \pm 431$ & $3958 \pm 219$
        & $4660 \pm 592$ & $99 \pm 31$
        & $2739 \pm 306$ & $2550 \pm 1162$
        & $-24.43 \pm 0.85$ & $-3.52 \pm 0.24$ \\
        \bottomrule
    \end{tabular}
    }
\end{table}

\begin{table}[t]
    \caption{Final dormant-neuron ratios (\%) on MuJoCo. Plasticity baselines generally suppress dormant neurons more strongly than reward inflation, but this does not consistently translate into better performance (Table~\ref{tab:plasticity_baselines}).}
    \label{tab:plasticity_dormancy}
    \centering
    \small
    \setlength{\tabcolsep}{3pt}
    \begin{tabular}{lrrrrrrrr}
        \toprule
        \textbf{Method}
        & \textbf{HalfCheetah} & \textbf{Walker2d}
        & \textbf{Humanoid} & \textbf{Swimmer}
        & \textbf{Hopper} & \textbf{Ant}
        & \textbf{Pusher} & \textbf{Reacher} \\
        \midrule
        0\%
        & 40.12 & 43.06 & 77.06 & 62.83
        & 44.60 & 53.59 & 73.17 & 76.26 \\
        2\%
        & 38.09 & 39.20 & 74.01 & 52.70
        & 49.83 & 38.46 & 72.42 & 76.17 \\
        10\%
        & 32.88 & 32.32 & 69.13 & 38.30
        & 43.15 & 36.89 & 68.42 & 78.02 \\
        \midrule
        ReDo
        & 12.43 & 8.81 & 34.55 & 27.20
        & 11.77 & 12.63 & 31.51 & 48.21 \\
        Reset
        & 18.24 & 19.60 & 71.35 & 51.35
        & 28.45 & 24.48 & 58.18 & 67.52 \\
        S\&P
        & 17.96 & 17.68 & 58.18 & 51.81
        & 23.62 & 26.74 & 54.35 & 67.99 \\
        \bottomrule
    \end{tabular}
\end{table}

\begin{table}[h!]
    \caption{Percentage of runs whose final greedy policy reaches the low-reward goal more often than the high-reward goal during evaluation. Higher exploration uses slower epsilon decay.}
    \centering
    \small
    \begin{tabular}{rcc}
        \toprule
        \textbf{Inflation} & \textbf{Low exploration} & \textbf{Higher exploration} \\
        \midrule
        0\%   & 18.8\%  & 0.0\%  \\
        1\%   & 15.6\%  & 0.0\%  \\
        5\%   & 31.2\%  & 0.0\%  \\
        10\%  & 28.1\%  & 0.0\%  \\
        50\%  & 93.8\%  & 59.4\% \\
        100\% & 100.0\% & 96.9\% \\
        \bottomrule
    \end{tabular}
    \label{tab:tabular_q_trap}
\end{table}

\clearpage

\subsection{Adaptation after a Task Switch}
\label{app:forward_backward}
\begin{figure*}
  \centering
  \includegraphics[width=0.6\linewidth]{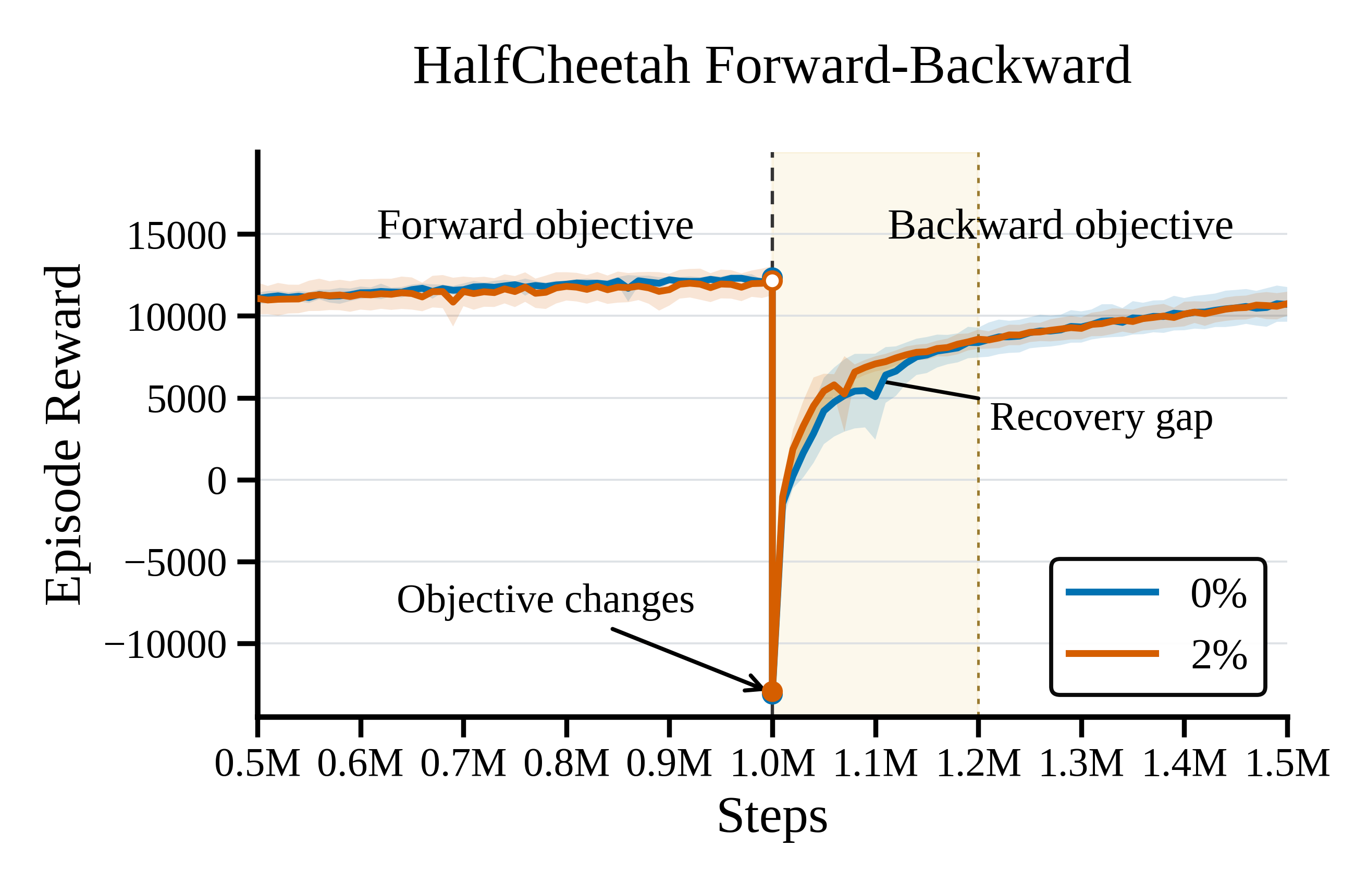}
  \caption{Evaluation episodic returns for SAC with 0\% and 2\% reward inflation (8 seeds per method). At 1M steps, the objective switches from forward to backward locomotion. Both the final forward-task and initial backward-task evaluations are shown at the switch. The highlighted region marks the early recovery period from 1.0M to 1.2M steps, during which 2\% inflation achieves higher mean returns.}
  \label{fig:continual_learning}
\end{figure*}
If the dormant-neuron suppression induced by reward inflation helps preserve plasticity, we would also expect benefits when learning successive tasks. To examine this possibility, we evaluate HalfCheetah agents by switching from forward to backward locomotion after 1M training steps, comparing 0\% and 2\% inflation across eight runs per condition. As shown in Figure~\ref{fig:continual_learning}, reward inflation leads to faster recovery following the objective change, while the performance of the two conditions becomes similar later in training. The 2\% condition achieved a higher post-switch AUC than the 0\% baseline (Welch's $t$-test, $p=0.044$). These results suggest a potential benefit for adaptation to changing objectives, although they do not establish dormant-neuron suppression as the cause of improved adaptation.

\subsection{Fed Hyperparameter Study}
\begin{figure*}
  \centering
  \includegraphics[width=1.0\linewidth]{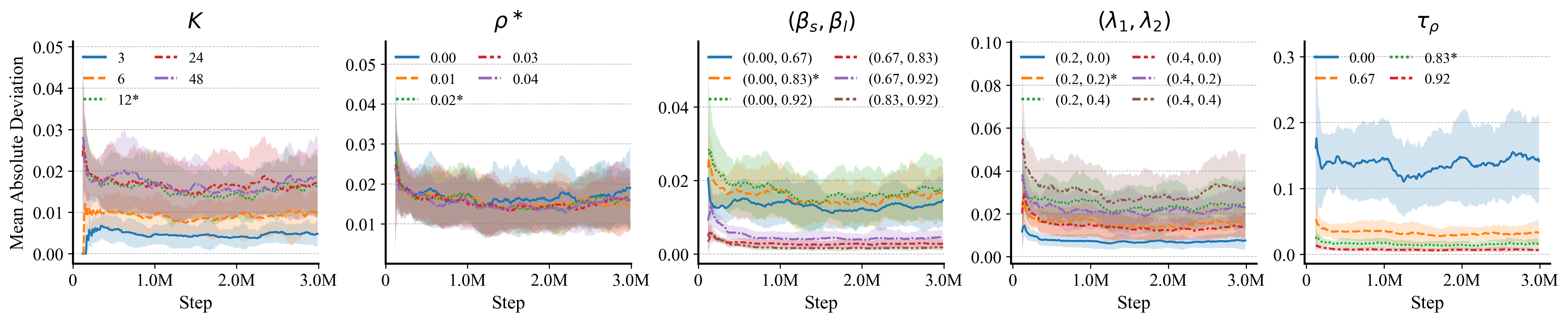}
  \caption{Mean absolute deviation curves of inflation rate under various Fed hyperparameter settings. Specifically, Larger $K$, $\lambda_1$, and $\lambda_2$, and smaller $\tau_\rho$ lead to more aggressive fluctuations in the inflation rate. For $\beta_s$ and $\beta_l$, smaller values and larger gaps between them amplify variations in gradient-norm measurements, which in turn induce greater inflation-rate volatility. Note that target inflation $\rho^\ast$ does not impact volatility.}
  \label{fig:federl_ablation_mad}
\end{figure*}

\begin{figure*}
  \centering
  \includegraphics[width=1.0\linewidth]{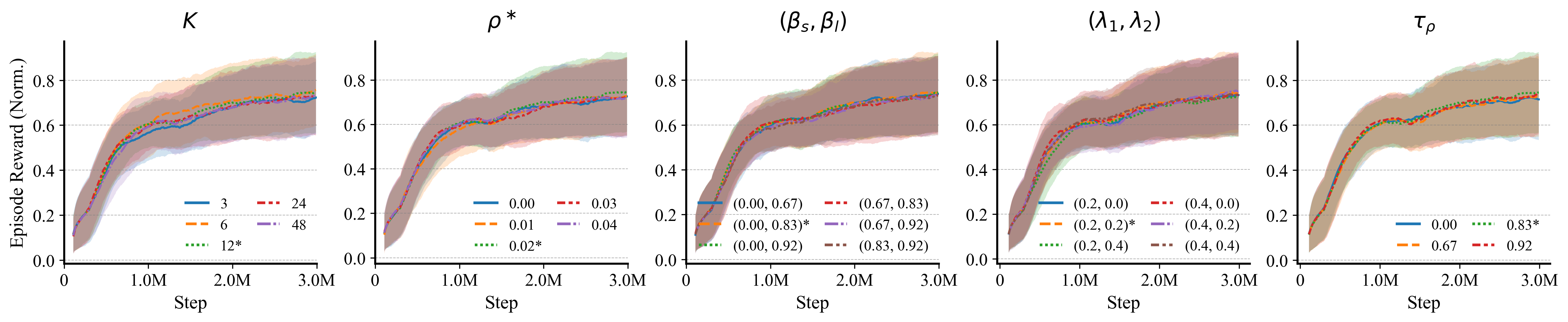}
  \caption{Learning curves of Fed varying 5 sets of hyperparameters. Default values are marked with $\ast$. Episode rewards are normalized across 4 tasks.}
  \label{fig:federl_hyper}
\end{figure*}

We conducted a hyperparameter study to assess the sensitivity of Fed to its hyperparameter choices.
Since the global horizon $Y$ mainly aids interpretation, while the effective inflation mechanism depends on the stepwise rate $\tilde{\rho}$, we fix $Y = 10^5$ and vary only the target inflation rate $\rho^\ast_Y$.
The remaining parameters control the volatility of inflation updates: larger $K$, $\lambda_1$, $\lambda_2$, and smaller $\tau_\rho$ increase fluctuations, while the relative scale of $\beta_s$ and $\beta_l$ affects the variability of gradient-norm measurements driving these updates.
Figure~\ref{fig:federl_ablation_mad} summarizes how these hyperparameters influence inflation-rate stability, while Figure~\ref{fig:federl_hyper} illustrates the sensitivity of normalized performance across four MuJoCo tasks when varying five representative hyperparameters and keeping the others fixed at their default settings.
Although some settings (e.g., $K = 6,12$ or $\rho^\ast = 0.2$) slightly improve performance, the overall variation remains relatively small.
This suggests that Fed is reasonably robust to moderate changes in hyperparameter settings.

\subsection{Environmental Uncertainty}
\begin{figure*}
  \centering
  \includegraphics[width=0.9\linewidth]{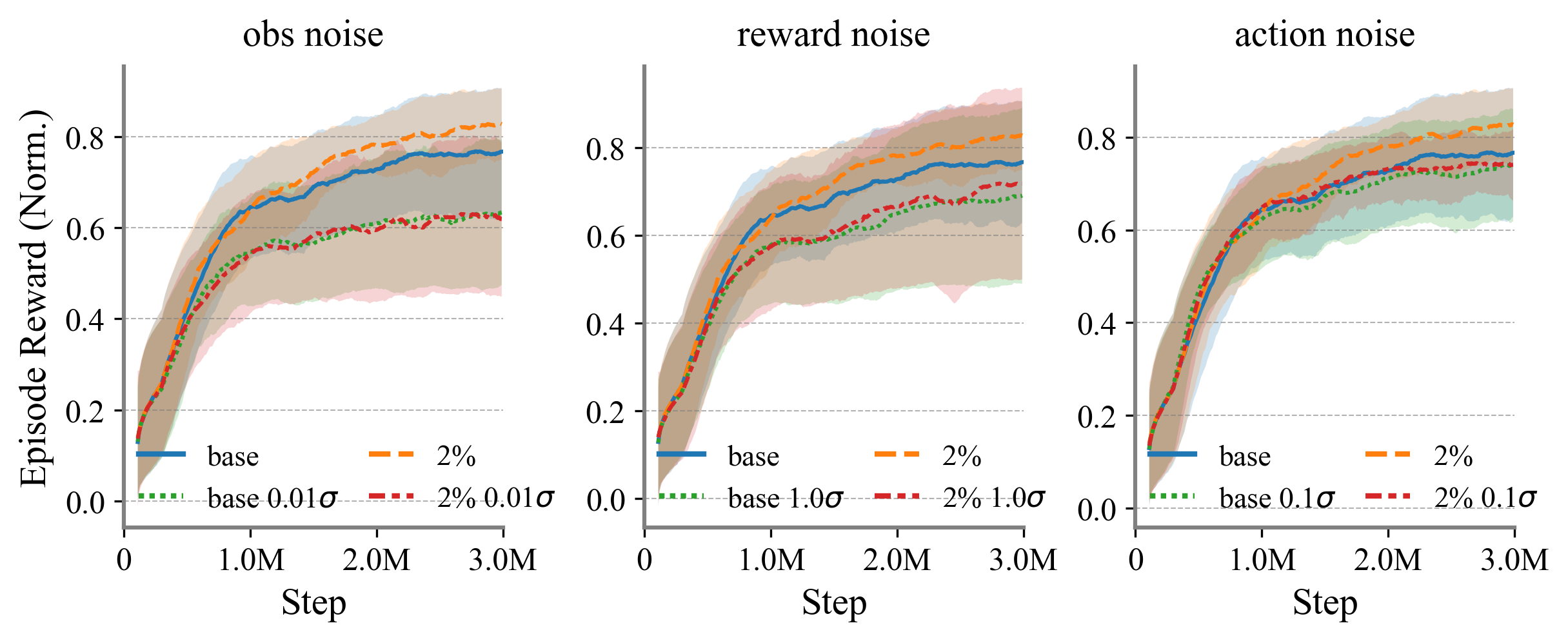}
  \caption{Non-inflation (0\%) vs. inflation (2\%) on more uncertain conditions.}
  \label{fig:noise}
\end{figure*}

Since reward inflation continuously changes the reward scale during training, we further investigated whether its benefits persist under additional environmental noise.
To examine this, we injected three types of Gaussian noise: (1) observation noise ($0.01\sigma$), (2) reward noise ($1.0\sigma$), and (3) action noise ($0.1\sigma$) into four MuJoCo tasks and compared SAC agents under $0\%$ and $2\%$ inflation.
As shown in Figure~\ref{fig:noise}, both agents experienced performance degradation under noisy conditions; however, the $2\%$ inflation agent maintained comparable or slightly better performance than the baseline.
These results suggest that the benefits of reward inflation can persist, at least partially, even when environmental noise is compounded with the inherent temporal inconsistency induced by inflation.

\clearpage

\section{Convergence Discussion}
\label{app:convergence}

Reward inflation introduces two related effects relevant to convergence: the overall reward scale changes during training, and the replay buffer contains transitions collected at different reward scales. We discuss how these effects behave under persistent positive inflation and when inflation stops. We consider discounted tabular Q-learning with bounded original rewards, assuming the standard convergence conditions on state-action coverage, diminishing learning rates, and replay sampling.

\paragraph{Changing reward scales.}
Let $g$ denote the current training step and $h\leq g$ the collection step of a transition in the replay buffer $\mathcal{D}_g$. Its stored reward is $I(h)r$. Under assumption (A1), the buffer-averaged multiplier
\[
    \bar I_g := \mathbb{E}_{\mathcal{D}_g}[I(h)]
\]
defines the approximate reference value function $\bar I_g Q^*$ in Eq.~(15). Thus, changes in $\bar I_g$ move the reference that the learner tracks.

Individual transitions also differ from this common reference scale. For the same transition and a fixed Q-function, the targets using $I(h)r$ and $\bar I_g r$ differ by
\[
    |y_{g,h}-\bar y_g|
    = |I(h)-\bar I_g|\,|r|
    \leq R_{\max}|I(h)-\bar I_g|,
\]
where $|r|\leq R_{\max}$. This captures the additional target variation arising from heterogeneous reward scales in replay.

\paragraph{Persistent positive inflation.}
Under a constant positive inflation rate, $I(g)$ grows exponentially. With a finite FIFO replay buffer, $\bar I_g$ follows this growth, so the reference $\bar I_g Q^*$ generally does not approach a finite limit. Transitions of different ages also retain different reward multipliers, with persistent relative differences and growing absolute differences. Consequently, convergence to a finite Q-function is generally not guaranteed. This does not imply that the greedy policy must diverge, since a common positive scaling preserves action rankings.

\paragraph{Stopping or decaying inflation.}
If inflation stops at a finite step $G$, the multiplier remains fixed at $I(G)$. Once earlier transitions leave the replay buffer, all stored rewards use this same multiplier. Both effects described above disappear, and learning reduces to ordinary Q-learning with uniformly scaled rewards. Under the stated convergence conditions, the Q-function converges to $I(G)Q^*$, which induces the same greedy optimal policies as $Q^*$.

The same limiting reward structure arises if inflation decays sufficiently rapidly that $I(g)\to I_\infty\in(0,\infty)$ and the buffer refreshes. In this case, convergence additionally requires the remaining scale perturbations to satisfy the relevant stochastic-approximation conditions. A summable sequence of nonnegative per-step inflation rates ensures a finite limiting multiplier; merely having the rate approach zero does not. Our fixed-inflation experiments examine finite-horizon learning benefits rather than such a convergence-oriented schedule.

\end{document}